%% file: main.tex
\documentclass{article}

\usepackage[preprint]{neurips_2026}

\usepackage[T1]{fontenc}
\usepackage[utf8]{inputenc}
\usepackage{microtype}
\usepackage{amsmath}
\usepackage{amssymb}
\usepackage{bm}
\usepackage{array} 
\usepackage{booktabs}
\usepackage{enumitem}
\usepackage{multirow}
\usepackage{graphicx}
\usepackage{subcaption}
\usepackage{xcolor}
\usepackage{url}
\usepackage{hyperref}
\usepackage{cleveref}
\usepackage[nolist]{acronym}

\setlist{nosep,leftmargin=*,topsep=3pt,partopsep=0pt,parsep=0pt,itemsep=1pt}
\makeatletter
\renewcommand\paragraph{\@startsection{paragraph}{4}{\z@}%
  {0.6ex plus 0.2ex minus 0.1ex}%
  {-0.8em}%
  {\normalfont\normalsize\bfseries}}
\makeatother

\graphicspath{{figure/}{../figures/}}

\title{GroupAffect-4: A Multimodal Dataset of Four-Person Collaborative Interaction}
\author{%
\begin{minipage}{0.98\textwidth}
\centering
\small
Meisam Jamshidi Seikavandi\textsuperscript{1,2},
Alice Modica\textsuperscript{1,3}\thanks{These authors contributed equally.},
Anna Obara\textsuperscript{1,4}\footnotemark[1],
Shan Ahmed Shaffi\textsuperscript{1},
Fabricio Batista Narcizo\textsuperscript{1,2},
Tanya Ignatenko\textsuperscript{1},
Ted Vucurevich\textsuperscript{1},
Karim Haddad\textsuperscript{1},
Daniel Barratt\textsuperscript{3},
Daniel Overholt\textsuperscript{4},
Jesper B{\"u}nsow Boldt\textsuperscript{1},
Paolo Burelli\textsuperscript{2},
Andrew Burke Dittberner\textsuperscript{1}
\\[0.75em]
\textsuperscript{1}GN Advanced Science, GN Group, Ballerup, Denmark\\
\textsuperscript{2}IT University of Copenhagen, brAIn lab, Copenhagen, Denmark\\
\textsuperscript{3}Copenhagen Business School, Copenhagen, Denmark\\
\textsuperscript{4}Aalborg University, Denmark
\end{minipage}
}

\begin{document}

\begin{acronym}
  \acro{BIDS}[BIDS]{Brain Imaging Data Structure}
  \acro{VFOA}[VFOA]{visual focus of attention}
  \acro{VAD}[VAD]{Valence-Arousal-Dominance}
  \acro{PAD}[PAD]{Pleasure-Arousal-Dominance}
  \acro{BFI}[BFI-44]{Big-Five Inventory}
  \acro{LSL}[LSL]{Lab Streaming Layer}
  \acro{SAM}[SAM]{Self-Assessment Manikin}
  \acro{PPG}[PPG]{photoplethysmography}
  \acro{EDA}[EDA]{electrodermal activity}
  \acro{IMU}[IMU]{inertial measurement unit}
  \acro{QC}[QC]{quality control}
  \acro{HRV}[HRV]{heart-rate variability}
  \acro{RMSSD}[RMSSD]{root mean square of successive differences}
  \acro{DUA}[DUA]{Data Use Agreement}
  \acro{FAIR}[FAIR]{Findable, Accessible, Interoperable, and Reusable}
  \acro{LOGOCV}[LOGO-CV]{leave-one-group-out cross-validation}
  \acro{KNN}[KNN]{$k$-nearest-neighbour}
  \acro{AUC}[AUC]{area under the curve}
  \acro{BHFDR}[BH-FDR]{Benjamini--Hochberg false discovery rate}
  \acro{MAE}[MAE]{mean absolute error}
  \acro{NGT}[NGT]{nominal group technique}
  \acro{XDF}[XDF]{Extensible Data Format}
  \acro{MAD}[MAD]{median absolute deviation}
  \acro{SSE}[SSE]{Server-Sent Events}
  \acro{HR}[HR]{heart rate}
  \acro{SCR}[SCR]{skin conductance response}
  \acro{RR}[RR]{R-R interval}
  \acro{ET}[ET]{eye tracking}
  \acro{EEG}[EEG]{electroencephalography}
  \acro{GSR}[GSR]{galvanic skin response}
  \acro{DBFS}[dBFS]{decibels relative to full scale}
\end{acronym}

\maketitle

\begin{abstract}
Existing affective-computing, social-signal-processing, and meeting corpora capture important parts of human interaction, but they rarely support analysis of affect in co-located groups as a coupled individual, interpersonal, and group-level process. The required signals (per-participant physiology, eye movement, audio, self-report, task outcomes, and personality) are usually fragmented across separate dataset traditions.
We introduce \textbf{GroupAffect-4}, a multimodal corpus of 40 participants in 10 four-person groups, each completing four ecologically varied collaborative tasks spanning information pooling, negotiation, idea generation, and a public-goods game.
Each participant is instrumented with a wrist-worn physiology sensor, eye-tracking glasses, and a close-talk microphone; sessions include continuous affect self-reports, post-task questionnaires, task outcomes, and Big-Five personality scores, all time-aligned to a shared clock.
The dataset covers over 91\% of expected physiology windows and 98\% of eye-tracking windows, with strong task validity confirmed by a clear affective manipulation check across the negotiation block.
We define fifteen benchmarkable targets spanning three analysis levels within-person state, between-person traits, and group dynamics and report leave-one-group-out feasibility baselines establishing the dataset's evaluative scope.
GroupAffect-4 is released with a \ac{BIDS}-inspired structure, Croissant metadata, a datasheet, per-session quality reports, and open processing scripts.
Code and processing scripts are available at \url{https://github.com/meisamjam/GroupAffect-4}; the dataset is publicly archived at \url{https://zenodo.org/records/20037847}.
\end{abstract}

\section{Introduction}
\label{sec:intro}

Meetings are a central setting where affect, cognition, and social dynamics co-occur under collaborative pressure \cite{carletta2005ami,janin2003icsi,marks2001teamprocesses}.
Understanding them requires observation at three co-occurring levels: within each person (physiological state, attention, cognitive load), between people (personality, trust, negotiation stance), and across the group (floor balance, coordination, collective outcomes), following multilevel views of team process and social interaction \cite{marks2001teamprocesses,vinciarelli2009socialsignalprocessing}. Yet, most multimodal affect corpora address at most one of these levels, and typically in single-participant or dyadic settings \cite{ringeval2013recola,busso2008iemocap,koelstra2012deap}.
Co-located groups impose qualitatively different constraints: turn-taking is multi-party, gaze targets multiply, roles can shift within a session, and physiological responses must be interpreted against an evolving social context \cite{sacks1974turntaking,goodwin1981conversational,vertegaal2001gaze,vinciarelli2009socialsignalprocessing}.
Theoretically, this situates group affect at the intersection of individual appraisal processes, where personal relevance and coping capacity shape felt response \cite{lazarus1991emotion,scherer2001appraisal}, and interpersonal emotion dynamics, where the ongoing reactions of co-present others can reshape appraisal, expression, and regulation \cite{parkinson1996emotions_social,vankleef2009emotion_social_information,barsade2002emotionalcontagion}.
Datasets combining per-participant physiology, gaze, audio, task outcomes, self-reports, and personality measures in the same co-located group setting are rare among the publicly described corpora reviewed in \Cref{sec:dataset_positioning,tab:dataset_comparison}.

\begin{figure*}[t]
  \centering
  \includegraphics[width=\linewidth]{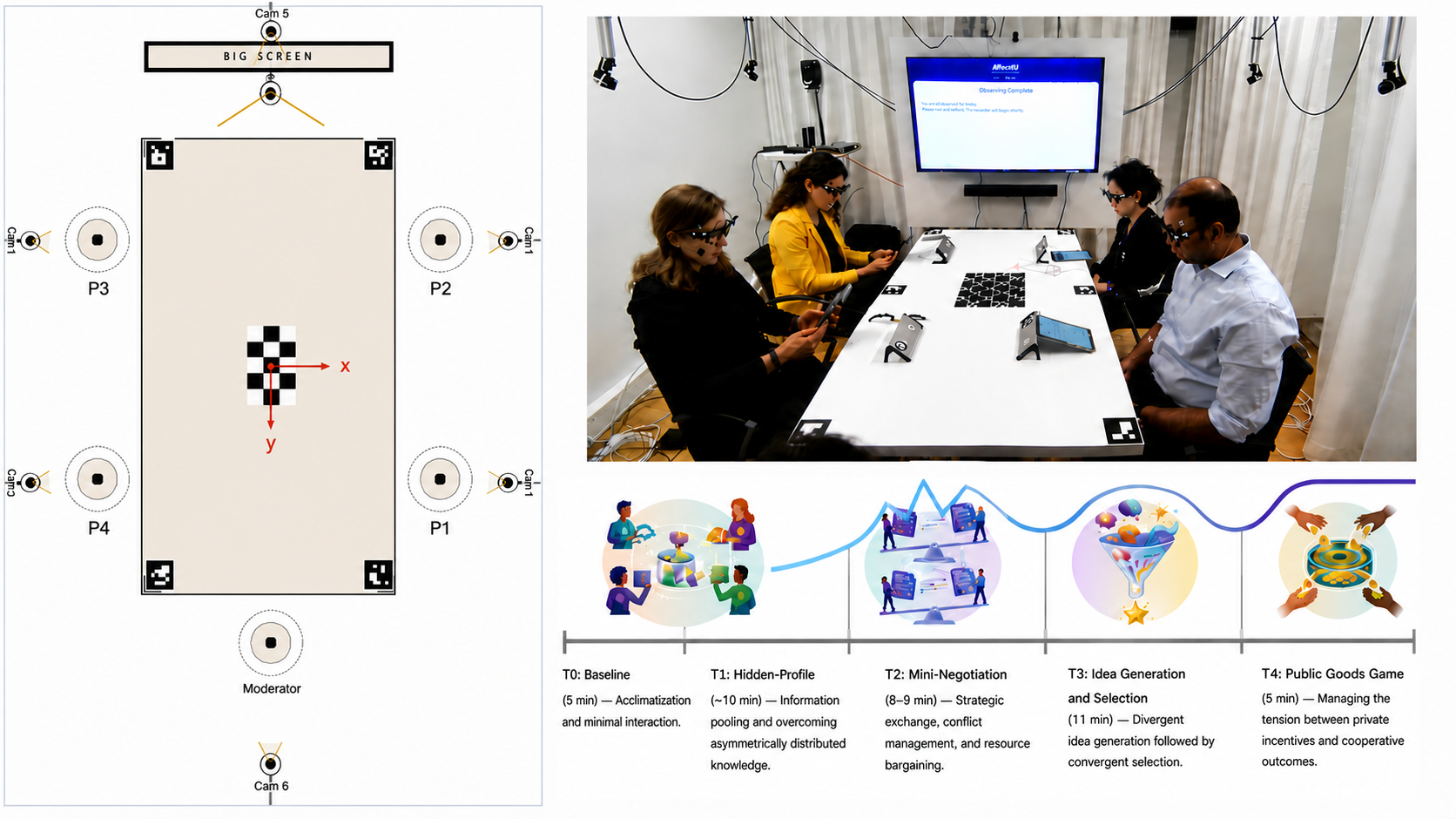}
  \caption{GroupAffect-4 study design and release structure.
    \textbf{Left:} top-down lab layout with four co-located participants (P1–-P4) seats around a 1.80 m × 0.80 m × 0.75 m rectangular table, seven camera positions, and ArUco calibration markers. 
    \textbf{Upper Right:} participants during a task session wearing Tobii Glasses 3, EmotiBit sensors and lapel microphones; tablets collect self-reported SAM valence, arousal, and dominance; a shared screen displays task prompts. 
    \textbf{Lower Right:} session timeline showing baseline (T0) followed by four collaborative tasks (T1: hidden-profile decision; T2: mini-negotiation; T3: idea generation; T4: public-goods game), with \ac{VAD} probes collected at intervals.}
  \label{fig:overview}
\end{figure*}

We present \textbf{GroupAffect-4}, a multimodal corpus of four-person group interaction designed to support analysis at three levels: within-person affective and cognitive state, between-person stable traits, and group-level interaction dynamics within the same synchronised release.
The motivating premise is not scale, but density: each participant is observed simultaneously through autonomic sensing, egocentric eye tracking, close-talk audio, task outcomes, and in-situ self-report, so the same session can support questions about regulation, attention, conversational floor dynamics, and
cooperation that are usually split across separate datasets.
Simultaneity matters because cross-level relationships such as whether physiological arousal tracks subjective dominance, or whether audio overlap predicts cognitive demand independently of physiology, can only be tested when modalities are co-observed at the same moment across all participants.
This design also builds on prior work showing that gaze dynamics, personality, and multimodal physiological signals can contribute complementary information
for emotion perception and face-to-face affect modelling
\cite{seikavandi2023gaze,seikavandi2025affec,seikavandi2025interplay,seikavandi2025mumtaffect}.
This paper is a dataset-characterisation paper: its primary contribution is a transparent description of what the dataset contains, how modalities are
synchronised, what can be evaluated, and where the main limitations lie.
\Cref{fig:overview} gives a schematic of the study design, modality stack, task sequence, and benchmarkable outputs.

\section{Literature and Dataset Positioning}
\label{sec:dataset_positioning}

GroupAffect-4 is a compact, high-density dataset rather than a scale-first corpus, following the dataset-characterisation tradition in which value depends not only on sample size but also on documentation, synchronisation, annotation quality, and evaluative scope \cite{gebru2021datasheets,wilkinson2016_fair,neurips2026ed}.
Studying group affect requires capturing three co-occurring levels simultaneously: individual affective and cognitive state, interpersonal dynamics, and group-level coordination and outcomes \cite{marks2001teamprocesses,vinciarelli2009socialsignalprocessing}.
Individual state is supported by physiology, pupil, and self-report; interpersonal dynamics by audio and gaze; and group-level coordination by floor-balance and outcome measures.
The central comparison question is therefore not which prior corpus is largest, but which combines the per-participant signals needed at each level -- wearable physiology, egocentric gaze, close-talk audio, in-situ self-report, explicit task outcomes, and personality measures -- in one synchronised co-located four-person release.
Classic affective corpora provide necessary background: IEMOCAP
\cite{busso2008iemocap}, RECOLA \cite{ringeval2013recola}, and DEAP
\cite{koelstra2012deap} address dyadic or individual settings; AMIGOS adds
group media-viewing with wearable physiology and personality
\cite{miranda2021amigos}; K-EmoCon provides wearable sensing with rich
emotion annotation in dyadic debates \cite{park2020kemocon}; UDIVA covers
dyadic interaction with personality and physiology \cite{palmero2021udiva}.
Among multiparty corpora, AMI and ICSI are canonical meeting baselines
\cite{carletta2005ami,janin2003icsi}; ELEA, GAP, and UGI add small-group
decision tasks with questionnaires and performance measures
\cite{sanchez2011elea,zhang2018gap,bhattacharya2019ugi}.
The KTH-Idiap Group-Interviewing corpus provides four-person discussions
with close-talk microphones and gaze/\ac{VFOA} design \cite{muller2018kthidiap};
GaMMA contributes four-person polyadic conversations with eye-tracking
glasses, motion, and multi-talker audio \cite{dourado2026gamma};
a recent agile-team dataset covers four-person software teams with audio,
visual attention, and equity surveys \cite{agileteam2025dataset};
and G-REx contributes large-scale group PPG/EDA in naturalistic movie
sessions \cite{martinez2023grex}.

The ingredients already exist but are distributed across corpora: task
structure and outcomes in AMI, ELEA, GAP, UGI, and the agile-team dataset;
physiology and affect labels in AMIGOS, K-EmoCon, and G-REx; gaze-rich
multiparty sensing in KTH-Idiap and GaMMA.
\Cref{tab:dataset_comparison} tabulates the closest comparators per axis.
Among publicly described datasets reviewed here, we did not find an earlier
dataset that clearly combines all these strands in one co-located four-person
benchmarkable release.
GroupAffect-4 is positioned between fully lab-controlled paradigms (tightly constrained illumination, timing, and stimuli) and fully naturalistic settings (uncontrolled context), reflecting the broader tension in affective computing between experimental control, ecological validity, and annotation reliability \cite{zeng2009survey,schuller2011speech_emotion,vinciarelli2009socialsignalprocessing}: structured \emph{task-context-primed} windows and LSL synchronisation provide auditability, while unscripted group interaction within those windows preserves ecological realism.

\begin{table*}[t]
\centering\footnotesize
\setlength{\tabcolsep}{4pt}
\caption{Closest dataset comparators for GroupAffect-4, organised by nearest shared axis.
  \textbf{Grp.}~=~group/dyad size.
  Feature columns: \textbf{Ph}~=~wearable physiology (EDA/PPG),
  \textbf{ET}~=~mobile egocentric eye-tracking,
  \textbf{Au}~=~per-participant close-talk audio,
  \textbf{SR}~=~in-situ affect self-report,
  \textbf{Pe}~=~personality scores,
  \textbf{Ta}~=~structured collaborative tasks with outcomes.
  \checkmark~present;\enspace$\circ$~partial/limited;\enspace---~absent.
  GroupAffect-4 (\textbf{this work}) provides all six.}
\label{tab:dataset_comparison}
\begin{tabular}{@{}p{0.16\linewidth}cp{0.31\linewidth}cccccc@{}}
\toprule
\textbf{Dataset} & \textbf{Grp.} & \textbf{Nearest shared axis} &
\textbf{Ph} & \textbf{ET} & \textbf{Au} & \textbf{SR} & \textbf{Pe} & \textbf{Ta} \\
\midrule
AMI \cite{carletta2005ami} & 4 &
  4-person meetings; synchronised multimodal capture; rich annotations &
  --- & --- & $\circ$ & --- & --- & \checkmark \\
ELEA \cite{sanchez2011elea} & 4--5 &
  Small-group survival task; questionnaires, personality, performance &
  --- & --- & $\circ$ & \checkmark & \checkmark & \checkmark \\
GAP \cite{zhang2018gap} & small &
  Group decision task; transcripts and explicit performance outcomes &
  --- & --- & $\circ$ & \checkmark & --- & \checkmark \\
UGI \cite{bhattacharya2019ugi} & small &
  Group task; questionnaires; privacy-preserving multimodal capture &
  --- & --- & $\circ$ & \checkmark & $\circ$ & \checkmark \\
KTH-Idiap \cite{muller2018kthidiap} & 4 &
  4-person discussions; close-talk mics; \ac{VFOA} design &
  --- & $\circ$ & \checkmark & --- & --- & $\circ$ \\
MatchNMingle \cite{cabrera2018matchnmingle} & var. &
  Real-world social sensing; wearables, personality, group formations &
  $\circ$ & --- & $\circ$ & --- & \checkmark & --- \\
AMIGOS \cite{miranda2021amigos} & ind./grp &
  Wearable physiology, self-assessment, and personality &
  \checkmark & --- & --- & \checkmark & \checkmark & --- \\
K-EmoCon \cite{park2020kemocon} & 2 &
  Wearable sensing; rich emotion labels; dyadic social interaction &
  \checkmark & --- & \checkmark & \checkmark & $\circ$ & $\circ$ \\
G-REx \cite{martinez2023grex} & grp. &
  Large-scale group physiology in naturalistic (audience) settings &
  \checkmark & --- & --- & --- & --- & --- \\
GaMMA \cite{dourado2026gamma} & 4 &
  4-person conversations; ET glasses; multi-talker audio &
  --- & \checkmark & \checkmark & --- & --- & --- \\
Agile-team \cite{agileteam2025dataset} & 4 &
  4-person teams; audio, visual attention, task outcomes, equity surveys &
  --- & $\circ$ & \checkmark & \checkmark & $\circ$ & \checkmark \\
AFFEC \cite{seikavandi2025affec} & 2 &
  Face-to-face; EEG, ET, GSR, self-reports, personality &
  \checkmark & \checkmark & $\circ$ & \checkmark & \checkmark & --- \\
\midrule
\textbf{GroupAffect-4} & \textbf{4} &
  \textbf{This work} &
  \checkmark & \checkmark & \checkmark & \checkmark & \checkmark & \checkmark \\
\bottomrule
\end{tabular}
\end{table*}

\section{Dataset Design}
\label{sec:design}

\paragraph{Participants.}
Sixteen sessions were recorded in total: the first five were pilot sessions used to finalise the protocol and are not included in the release; of the
remaining 11 non-pilot sessions, 10 are released (one was excluded due to
incomplete modality coverage).
The final subset therefore contains 10 final groups of four participants each ($N_p=40$ unique participants), all with complete demographics and \ac{BFI} scores.
Mean age is 35.4 years (SD~9.2, range 23--58); 21 female and 19 male (demographics in \Cref{fig:demographics}).
Collected demographic fields include age, sex, handedness, English
proficiency, and education level; racial or ethnic composition was not
collected as part of the study protocol.
Participants gave written informed consent under a participant information statement reviewed and approved by the GN Hearing A/S Legal Department.
Within-session roles use anonymous seat identifiers P1--P4; the identity
mapping is not part of the release.
Pre-session familiarity ratings cluster at 1--3 on a 7-point scale
(cross-participant mean 1.68, SD\,1.60 for other-group-member ratings;
maximum group mean 3.08/7), indicating that most participants were meeting for the first time and that established social ties are unlikely to be a primary driver of the observed group dynamics.

\paragraph{Pre-session questionnaire.}
Before each session, participants completed the \ac{BFI} \cite{john1999bigfive}.
Per-domain scores are computed as item means after reverse coding;
all 40 participants have complete BFI-44 records.

\paragraph{Session structure.}
Each session ($\approx$70~min) comprised a free-talk social baseline (T0), in which participants engaged in unstructured conversation to allow
familiarisation and sensor stabilisation, a common concern in psychophysiological recording and wearable affect measurement \cite{taskforce1996hrv,posadaquintero2020_eda_review,mathot2018_pupillometry_review}, followed by four collaborative tasks, with task onsets and offsets demarcated by \ac{LSL} markers and written to release-side task-window files.
All sessions followed the same fixed T0$\to$T1$\to$T2$\to$T3$\to$T4 sequence with no counterbalancing. Any cross-task effects should therefore be interpreted as effects within this fixed task order, where fatigue, and rapport also vary, rather than as effects that can be attributed solely to the tasks themselves.

\paragraph{Tasks.}
The four tasks were selected to target distinct social–affective regimes (\Cref{tab:tasks}): T1 emphasizes information pooling, group reasoning, and consensus formation under initially asymmetric information \cite{stasser1985_hiddenprofile,lu2012_hiddenprofile_meta}; T2 introduces interdependent interests and negotiation pressure, giving rise to concession patterns, cooperative negotiation strategies, and integrative trade‑offs \cite{vankleef2004_emotion_negotiation,vankleef2009emotion_social_information}; T3 focuses on creative idea generation, engagement in group discussion, and convergence toward a shared solution\cite{rietzschel2010selection}; and T4 uses a small public‑goods game in which private contributions determine individual and group payoffs, highlighting cooperative tension, fairness evaluations, and social comparison \cite{fehr2000_publicgoods_punishment,chaudhuri2011_publicgoods_survey}.
Key outcomes: T1 group consensus rate 100\% (above the $\approx$20--40\% literature baseline \cite{lu2012_hiddenprofile_meta}); T2 all groups settled but overran the 8-min guideline (mean 11.4\,min, SD\,1.6); T3 winning-idea records recovered for 10/10 groups; T4 mean contribution 6.92/10 tokens (SD\,3.10), higher than typical one-shot rates \cite{chaudhuri2011_publicgoods_survey}, likely reflecting rapport built across T1--T3.

\paragraph{Self-report probes.}
Valence, arousal, and dominance probes are collected at scheduled moments during the tasks - excluding dominance for T4 - on a 9-point \ac{SAM} scale, following dimensional affect and \ac{PAD}/\ac{VAD} measurement traditions \cite{mehrabian1974pad,russell1980circumplex,bradley1994sam}.
In-task items probe each participant's first-person \emph{felt} state \cite{russell1980circumplex,bradley1994sam}. Each task is followed by a brief post-task questionnaire that targets participants' evaluations of the task and of their interactions with the group. They use 1--7 Likert scales for both self-directed evaluations, and other-directed \emph{perceived} appraisals of group members’ influence and relational stance, captured via per-seat dominance and trust ratings.
\cite{vankleef2009emotion_social_information,parkinson1996emotions_social}.
In-task probes are administered as retrospective snapshots at scheduled phase boundaries, with near-complete coverage and per-task completeness (98.8\% overall), as documented in \Cref{app:stimuli}.

\input{tables/task_overview}

\section{Modalities and Acquisition}
\label{sec:modalities}

The GroupAffect-4 version~1.0 release (no video) contains synchronised physiology, egocentric
gaze/pupil features, audio-derived features, transcript artifacts,
behavioural responses, task outcomes, and personality metadata.
Multi-camera room video and scene video were recorded but are deferred to a future release pending full quality-control.
The modalities are listed in \Cref{tab:modalities} (see \Cref{app:stimuli}) and are designed as complementary observational channels rather than redundant measures of a single latent construct, following multimodal affective-computing and social-signal-processing practice \cite{vinciarelli2009socialsignalprocessing,zeng2009survey,koelstra2012deap}.
Audio captures conversational floor dynamics and prosodic variation core behavioural channels in social interaction and affective computing \cite{sacks1974turntaking,vinciarelli2009socialsignalprocessing,schuller2011speech_emotion}.
Pupil diameter indexes attentional engagement and cognitive load under natural
viewing conditions \cite{mathot2018_pupillometry_review}, while prior
face-viewing work suggests that gaze dynamics can carry emotion-perception
signal in naturalistic affective settings \cite{seikavandi2023gaze,seikavandi2025interplay}.
Wrist physiology (EDA, PPG, skin temperature) provides autonomic arousal proxies that operate largely independently of the social-behavioural channel \cite{posadaquintero2020_eda_review}.
Self-report supplies the subjective grounding that links observed signals to experienced states, while also introducing the usual limitations of retrospective and probe-based affect measurement \cite{russell1980circumplex,bradley1994sam}.
The near-zero cross-modal correlation between pupil and physiology features ($|r|<0.10$ across 136 participant-task rows; \Cref{sec:stats}) suggests that these channels provide partly distinct information rather than redundant readouts motivating multimodal rather than unimodal benchmarks \cite{mathot2018_pupillometry_review,posadaquintero2020_eda_review}.

\paragraph{Physiology.}
Each participant wears an EmotiBit wrist sensor (\ac{PPG}, \ac{EDA}, skin temperature, \ac{IMU} at $\approx$25~Hz).
\ac{QC} pass: 182/200 rows observed (91.0\%); usable rates 76.0\% \ac{PPG}/\ac{EDA}/Temp, 90.5\% \ac{IMU} (\Cref{tab:stats}).
\ac{PPG}-derived \ac{HRV} should be interpreted cautiously given the 25~Hz rate ($\approx$40~ms \ac{RMSSD} quantisation floor) and sensitivity to motion artefacts \cite{taskforce1996hrv,shaffer2017hrv,georgiou2018ppghrv}.
EDA rows with elevated wrist motion are flagged \texttt{motion\_contaminated} rather than discarded; accelerometer data are retained as a companion QC signal \cite{taylor2015automaticeda,posadaquintero2020_eda_review}.

\paragraph{Eye tracking.}
Each participant wears a Tobii Pro Glasses~3 unit providing head-relative scene-frame gaze, pupil diameter, and validity flags at 50~Hz.
196/200 rows observed (98.0\%); 87.2\% gaze and 95.4\% pupil usability.
Room-frame gaze alignment is deferred to v2.
The release does not include luminance-normalised pupil modelling; pupil effects should be interpreted with residual light-reflex confounding in mind \cite{mathot2018_pupillometry_review}.

\paragraph{Audio.}
Per-participant close-talk DPA~4060 microphones are recorded at 48~kHz, 24~bit via an RME Fireface~802 interface.
The public release includes prosodic features (opensmile GeMAPSv01b), transcript artifacts, speech-activity flags, and interaction metrics (speaking fraction, pause count, overlap).
Audio quality control achieved a 95\% pass rate (38/40 participant-task rows usable; two participants flagged for minimal vocal activity).
Raw audio is gated under a \ac{DUA} due to voice re-identification risk \cite{nautsch2019preserving,tomashenko2021voiceprivacy}.

\paragraph{Synchronisation.}
Acquisition is organised around a common \ac{LSL} timebase \cite{kothe2025_lsl}.
Redundant timing anchors are preserved so offsets, missing streams, and
alignment assumptions can be inspected post-processing.
For AV devices, frame-log start-spread analyses target sub-10~ms alignment
as a dataset-level QC criterion.
For the DPA~4060 microphones, the hardware clock drifts by approximately 0.04~ms/s relative
to the \ac{XDF}/\ac{LSL} clock, so task clipping uses per-microphone linear anchor
correction rather than a single global offset; in documented sessions this
reduces practical task-level residual sync error to below 5~ms
(\Cref{app:quality}).

\section{Processing and Release}
\label{sec:processing}

The release follows a \ac{BIDS}-inspired structure \cite{gorgolewski2016brain} organised around \ac{FAIR} principles \cite{wilkinson2016_fair}, in which original neuroimaging-focused BIDS conventions are adapted to a multimodal setting that combines physiology, eye tracking, audio-derived features, behavioural responses, and task metadata.
Per-modality subdirectories hold physiology, eye tracking, audio-derived features, transcripts, behavioural responses, and annotations; the processing pipeline is deterministic and reproducible.
Feature tables contain task-window summary statistics (means, delta-from-baseline values, prosodic summaries, turn-taking summaries).
For benchmarks (\Cref{sec:benchmarks}) a five-step modality-aware pipeline is applied: \ac{ET} quality gating, physiological plausibility gating, $\pm3\,\sigma$ winsorisation, within-person robust z-score normalisation (median/MAD across T1--T4), and fold-internal KNN imputation; full details are in \Cref{app:preprocessing}.
B0--B3d numbers should be read as biased characterisation estimates: within-person normalisation is computed before the LOGO split, creating a mild dependency between test-row statistics and training normalisation, a known source of optimistic bias in cross-validation when preprocessing is not fully nested within the training fold \cite{varma2006biascv,cawley2010overfitting}.
A leakage-free protocol, fit from training participants, and applied to held-out set, is recommended for future work.
Tabular modalities are released under CC~BY~4.0; raw audio is gated under a Data Use Agreement due to re-identification risk.
The release is documented with Croissant metadata, Responsible AI fields, and a datasheet, following current dataset-documentation and metadata practices \cite{gebru2021datasheets,akhtar2024croissant,mlcommons2024croissant,neurips2026hosting}.

\section{Benchmark Tasks and Feasibility Baselines}
\label{sec:benchmarks}

We define evaluation targets across three analysis levels and report feasibility baselines using
\ac{LOGOCV} (split key = \texttt{group\_id}), ensuring no participant in the test fold shares an interaction context with training data; the mild preprocessing leakage from within-person normalisation described in \Cref{sec:processing} applies independently of fold structure \cite{varma2006biascv,cawley2010overfitting}.
Baselines use a single Ridge regressor or logistic classifier with the same preprocessing pipeline as in \Cref{sec:processing} but no temporal modelling or architecture
search.
Within-person z-score normalisation is applied for within-person state
targets (B0--B3d); raw delta-from-baseline features (physiology and pupil) plus absolute
audio features are used for between-person targets (B4a--B5). Note that audio is baseline-normalised for B0--B3d (within-person z-score
across T1--T4) but kept in absolute form for B4a--B5, because T0 free-talk
audio is insufficiently controlled for a per-participant baseline; this
modality asymmetry should be kept in mind when comparing modality
contributions across benchmarks.
Residual missing values are imputed via \ac{KNN} ($k=5$) inside each fold
\cite{troyanskaya2001knn}.
The benchmarks are organised across three analysis levels: Level~1 within-person state (participant-task unit, within-person z-score normalisation), Level~2 between-person stable traits (participant unit, sample-level features), and Level~3 group dynamics (group-task unit, group-aggregated features); the normalisation rationale for each level is given above.
The separation between within-person state targets and between-person trait
targets follows prior personality-aware affect modelling work showing that
trait information can support affect inference but requires evaluation choices
that preserve inter-individual variation
\cite{john1999bigfive,seikavandi2025interplay,seikavandi2025mumtaffect}.
Four additions to the original B0--B7 suite are introduced here (marked $\dagger$): B3c and B3d extend Level~1 with social-affective constructs available in the existing survey data; B4c adds BFI Agreeableness to Level~2; and B6b tests whether within-group dispersion features resolve the B6a mean/variance mismatch.

\input{tables/benchmark_results}

\paragraph{Level~1   Within-person state.}
B3a mental demand is the clearest above-chance signal (\ac{AUC}~0.719; \Cref{tab:benchmark_results}), driven by audio overlap fraction, which we interpret as a proxy for conversational-floor competition \cite{sacks1974turntaking,vinciarelli2009socialsignalprocessing}.
A single audio feature surpasses the full 31-feature model for B3a, confirming that cognitive-load detection in group tasks is primarily a speech-floor phenomenon.
B1a valence is moderate (AUC~0.657) and B3c satisfaction is above chance (AUC~0.571); B1b arousal and B2 dominance are near chance, reflecting temporal label--signal mismatch and feature-set limitations respectively.
B3d trust pooled is near chance but rises to AUC~0.679 when restricted to T4 (cooperative context), where within-group trust variance is sufficient for fold-local splits to be meaningful.

\paragraph{Level~2   Between-person traits (challenge).}
LOGO-CV AUC for all three personality benchmarks (B4a--B4c) is near or below chance, driven by test folds of only four participants too small for stable AUC estimates regardless of true signal strength, especially when model selection and evaluation are performed under small grouped folds \cite{varma2006biascv,cawley2010overfitting}.
Spearman correlations on the full sample suggest that trait-related associations may be present (pupil$\times$Agreeableness $r=0.51$, $p=0.008$; \Cref{app:bench_notes}); the B4 trio is therefore better treated as an explicitly unsolved challenge than as evidence of absent signal.
B5 contribution is similarly near-chance with high fold variance (AUC~0.429, SD~0.290).

\paragraph{Level~3   Group dynamics.}
B6a, B6b, and B7 all fall at or below the naive baseline, suggesting that task-window aggregates lose the turn-by-turn floor dynamics that govern speaking inequality in multiparty conversation \cite{sacks1974turntaking,goodwin1981conversational,vinciarelli2009socialsignalprocessing}.
B6b rules out feature-design as the explanation: an auxiliary binary classifier on the raw speaking-fraction SD achieves AUC~0.952, confirming the signal is definitively present but inaccessible to ridge regression at $n=28$ rows.
Per-benchmark interpretation notes, modality ablation, and feature-importance analysis are in \Cref{app:bench_notes}.
Sequential conversation benchmarks (next-speaker prediction, turn-taking, overlap onset) are defined in \Cref{app:extended_benchmarks} as future work.

\begin{figure*}[t]
  \centering
  \includegraphics[width=\linewidth]{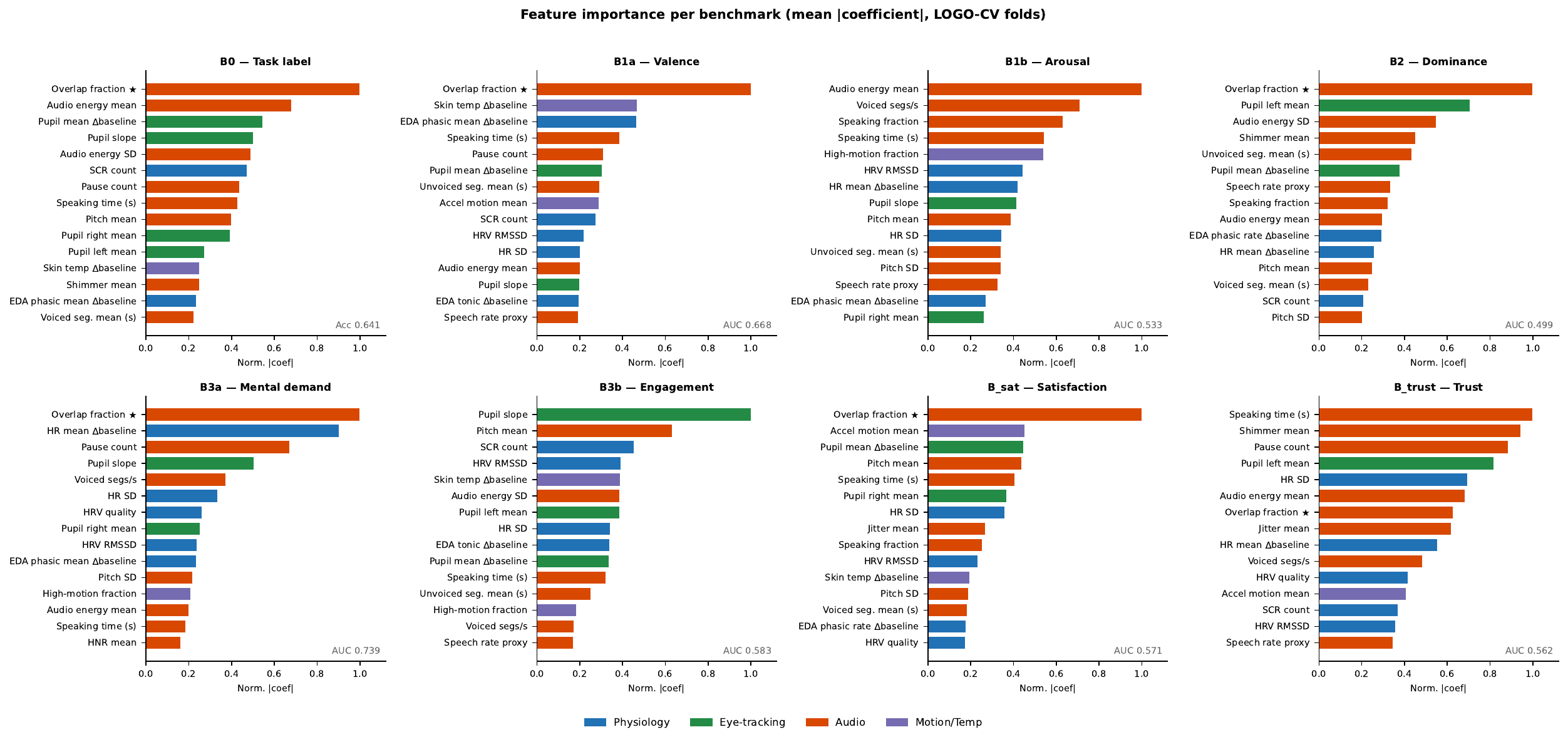}
  \caption{Per-benchmark ranked feature importance: top-15 features by mean
    normalised $|\text{coefficient}|$ across LOGO-CV folds (31-feature set).
    Bar colour: \textcolor[HTML]{2171b5}{\textbf{Physiology}},
    \textcolor[HTML]{238b45}{\textbf{Eye-tracking}},
    \textcolor[HTML]{d94801}{\textbf{Audio}},
    \textcolor[HTML]{756bb1}{\textbf{Motion/Temp}}.
    \textbf{$\star$}~=~\texttt{audio\_overlap\_fraction\_x}.
    B3a mental demand is dominated by a single audio-floor feature;
    B3b engagement loads instead on pupil slope and pitch, confirming a
    demand--engagement dissociation.
    Full benchmark notes and modality ablation in \Cref{app:bench_notes}.}
  \label{fig:feat_imp_ranked}
\end{figure*}

\section{Coverage, Quality, and Validation}
\label{sec:stats}

\Cref{tab:stats} reports modality coverage and task-response completeness; per-session detail is in \Cref{app:quality}.

\input{tables/dataset_stats}

\begin{figure}[t]
  \centering
  \includegraphics[width=\linewidth]{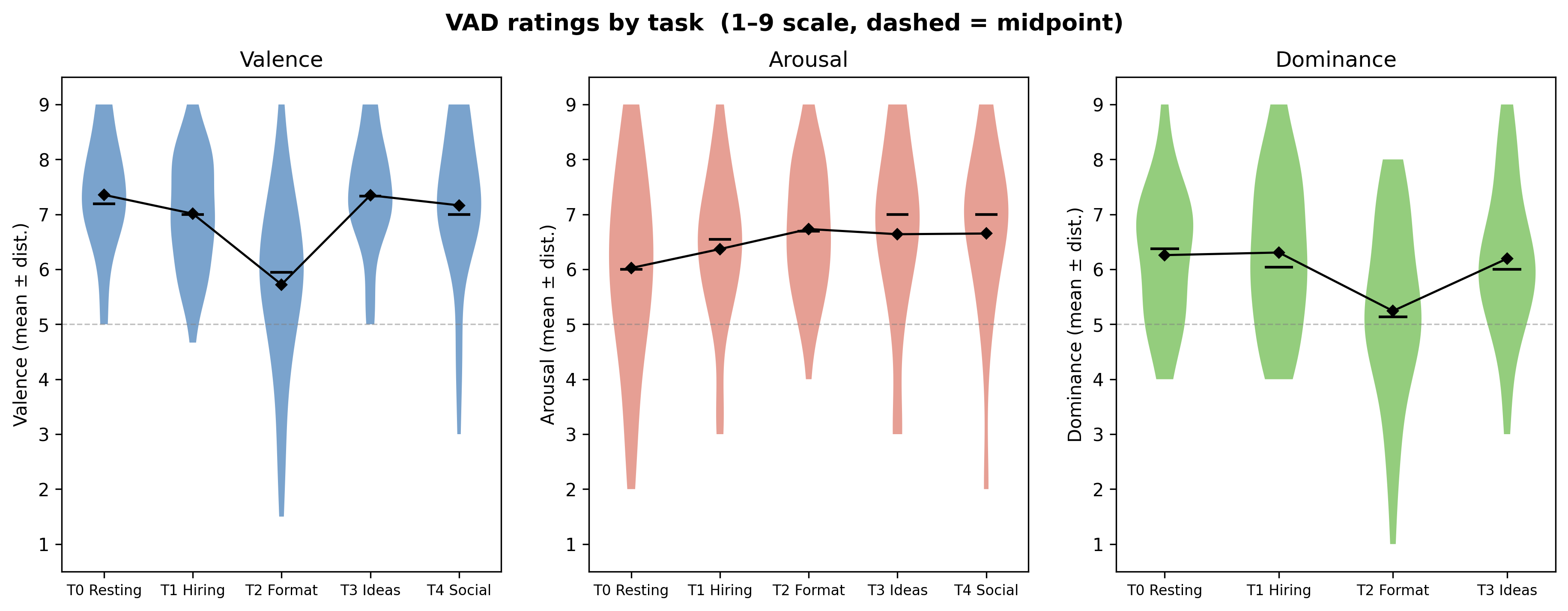}
  \caption{Valence, arousal, and dominance probe summaries by task. T2
    negotiation produces the largest valence drop ($d=1.06$)   the
    strongest manipulation check in the dataset. Dominance is absent for T4
    in the current response schema.}
  \label{fig:vad_by_task}
\end{figure}

\paragraph{Task validity.}
The key manipulation check is shown in \Cref{fig:vad_by_task}: mean valence drops sharply during T2 negotiation ($d=1.06$, $p=2.4\times10^{-9}$), and T2 overran its 8-min guideline in all 10 groups, indicating strong and consistent affective and behavioural differentiation.
Group-level trust remained above the scale midpoint after both T2 and T4, confirming collaborative rather than adversarial dynamics; the T2-to-T4 trust change is descriptive rather than statistically reliable at $n=10$ groups ($t(38)=0.96$, $p=0.34$).

\paragraph{Cross-modal complementarity and associations.}
Pupil and wearable physiology are weakly correlated ($|r|<0.10$), suggesting these channels provide partly distinct information rather than a redundant arousal readout an empirical argument for multimodal combination \cite{mathot2018_pupillometry_review,posadaquintero2020_eda_review}.
The strongest preprocessed associations are audio group-overlap fraction with post-block mental demand ($r=-0.62$, $q_{\text{FDR}}<0.001$, $n=99$) and pupil dilation with mental demand ($r=+0.36$, $q=0.008$, $n=99$), both motivating benchmark B3a \cite{sacks1974turntaking,mathot2018_pupillometry_review}.
Extended autonomic and conversation-structure analyses are in \Cref{app:worked_extended}.

\section{Discussion}
\label{sec:discussion}

The benchmark gradient across the three analysis levels is interpretable as a coherent finding about multimodal signal structure in collaborative meetings, not merely a performance ranking, consistent with the idea that different behavioural and physiological channels carry partly distinct social and affective information \cite{vinciarelli2009socialsignalprocessing,mathot2018_pupillometry_review,posadaquintero2020_eda_review}.
In brief: audio and pupil dominate cognitive-state and satisfaction detection; physiology contributes primarily to arousal-sensitive targets; personality prediction fails under the current normalisation and feature design, not because of signal absence; and group-level dynamics are not recoverable from task-window aggregates at this cohort size.

\paragraph{Mental demand and trust context.}
B3a mental demand is the clearest Level-1 signal (AUC~0.719), driven by a single audio overlap feature that exceeds the full 31-feature model \cite{sacks1974turntaking,vinciarelli2009socialsignalprocessing}.
B3c satisfaction shares the same audio-floor driver and is weaker.
B3d trust rises to AUC~0.679 when restricted to cooperative T4 context; T2's adversarial negotiation leaves near-zero within-group variance, making that fold-local split near-random, which motivates within-task trust probes in future releases \cite{parkinson1996emotions_social,vankleef2009emotion_social_information}.

\paragraph{Modality complementarity.}
The feature importance profiles in \Cref{fig:feat_imp_ranked} and modality ablation (\Cref{app:bench_notes}) together suggest audio and pupil provide largely non-redundant signal for cognitive-state detection, consistent with the near-zero physiology--pupil correlation ($|r|<0.10$; \Cref{sec:stats}) \cite{vinciarelli2009socialsignalprocessing,mathot2018_pupillometry_review,posadaquintero2020_eda_review}.
Physiology contributes primarily to arousal-sensitive targets via EDA phasic rate (B1b arousal, and to a lesser extent B1a valence) \cite{posadaquintero2020_eda_review}; B2 dominance is near chance.

\paragraph{Between-person inference.}
We interpret weak B4 results as reflecting the normalisation regime: within-person z-scoring removes inter-individual amplitude variance needed for between-person inference \cite{varma2006biascv,cawley2010overfitting}.
Future work should use fold-internal normalisation fitted on training participants only.

\paragraph{Group-level inequality.}
B6a, B6b, and B7 all fail the naive baseline.
B6b rules out feature-design: even within-group SD features fail (Ridge MAE~0.102 vs baseline~0.088), while raw SD binary classification reaches AUC~0.952 confirming the signal exists but task-window aggregates lose the turn-by-turn dynamics that govern floor inequality \cite{sacks1974turntaking,goodwin1981conversational,vinciarelli2009socialsignalprocessing}.
Per-turn voice-activity-detection extraction and a larger cohort ($n\geq20$ groups) are the prerequisites for this benchmark tier.

\section{Limitations}
\label{sec:limitations}

\paragraph{Sample size.}
40 participants across 10 groups supports dataset characterisation but not broad benchmark or leaderboard claims; LOGO-CV test folds of four participants make AUC estimates inherently inaccurate for participant-level targets.
\paragraph{Single site and language.}
All sessions were recorded at a single site and conducted in English; turn-taking norms and negotiation styles may reflect local linguistic, cultural, and institutional context \cite{sacks1974turntaking,goodwin1981conversational}.
\paragraph{Fixed task order.}
All sessions followed the same T0--T4 sequence with no counterbalancing; T4 cooperation outcome follows $\approx$60~min of prior collaboration and cannot be interpreted as a pure task effect without a counterbalanced replication.
\paragraph{Audio and modality scope.}
Raw audio is gated under a DUA due to re-identification risk \cite{nautsch2019preserving,tomashenko2021voiceprivacy}; lapel microphones can misclassify ambient noise; released gaze is in each participant's scene-camera frame and requires additional processing for cross-participant analysis.
Additional technical caveats (physiology precision, pupil comparability, survey ceiling, deferred modalities, no clinical ground truth) are in \Cref{app:extended_limitations}.

\section{Ethics, Access, and Responsible Use}
\label{sec:ethics}

The study was conducted under a participant information statement and informed consent procedure reviewed and approved by the GN Hearing A/S Legal Department.
Participants provided written informed consent covering all modalities.
Tabular features are released under CC~BY~4.0; raw audio is gated under a Data Use Agreement prohibiting re-identification, voice cloning, and non-research use \cite{nautsch2019preserving,tomashenko2021voiceprivacy}.
Released files use anonymous seat identifiers (P1--P4).
The dataset is not appropriate for clinical diagnosis, individual scoring, automated personnel evaluation, or surveillance \cite{gebru2021datasheets,mitchell2019modelcards}.
Detailed responsible-use guidance, disaggregated reporting expectations, and a 5-year maintenance commitment under a persistent-DOI repository are in the datasheet \cite{neurips2026hosting}.

\section{Conclusion}
\label{sec:conclusion}

GroupAffect-4 contributes a compact but unusually integrated resource for studying
co-located group affect: synchronised physiology, egocentric eye tracking,
audio, self-report, task outcomes, and personality measures from 10
four-person groups.
Its value lies in documentation, synchronisation, and multimodal density
around structured group tasks rather than scale.
The feasibility baselines show that the release supports meaningful evaluation
protocols while also exposing where current data are too small for broad
claims.
We release the dataset with Croissant metadata, a datasheet, explicit QC
tables, and conservative responsible-use guidance
\cite{gebru2021datasheets,akhtar2024croissant,mlcommons2024croissant}.


\clearpage
\bibliographystyle{plain}
\bibliography{references}

\newpage
\appendix

\section*{Appendix Table of Contents}

\small\noindent
\begin{tabular}{@{}lp{0.80\linewidth}@{}}
  \textbf{A} & \hyperref[app:acronyms]{List of Acronyms} \\
  \textbf{B} & \hyperref[app:stimuli]{Stimuli and Task Orchestration} \\
  \textbf{C} & \hyperref[app:extended_limitations]{Extended Limitations and Caveats} \\
  \textbf{D} & \hyperref[app:bfi44]{BFI-44 Scoring and Item List} \\
  \textbf{E} & \hyperref[app:audio_t0_baseline]{Audio T0 Baseline Reliability} \\
  \textbf{F} & \hyperref[app:sync]{Synchronisation Pipeline Detail} \\
  \textbf{G} & \hyperref[app:preprocessing]{Preprocessing Steps} \\
  \textbf{H} & \hyperref[app:worked_extended]{Extended Dataset Characterization} \\
  \textbf{I} & \hyperref[app:extended_benchmarks]{Extended Benchmarks: Sequential Conversation Tasks} \\
  \textbf{J} & \hyperref[app:bench_notes]{Benchmark Interpretation Notes} \\
  \textbf{K} & \hyperref[app:ablation]{Full Ablation Table} \\
  \textbf{L} & \hyperref[app:quality]{Per-Session Quality Table} \\
  \textbf{M} & \hyperref[app:rai]{Responsible AI and Croissant Metadata} \\
  \textbf{N} & \hyperref[app:datasheet]{Datasheet for Datasets} \\
\end{tabular}
\normalsize

\section{List of Acronyms}
\label{app:acronyms}

\begingroup
\footnotesize
\setlength{\tabcolsep}{5pt}
\begin{tabular}{@{}ll@{\quad}ll@{}}
  \toprule
  \textbf{Acronym} & \textbf{Expansion} & \textbf{Acronym} & \textbf{Expansion} \\
  \midrule
  AUC      & Area under the curve                        & LOGO-CV  & Leave-one-group-out cross-validation \\
  BFI-44   & Big-Five Inventory                          & LSL      & Lab Streaming Layer \\
  BH-FDR   & Benjamini--Hochberg false discovery rate    & MAD      & Median absolute deviation \\
  BIDS     & Brain Imaging Data Structure                & MAE      & Mean absolute error \\
  dBFS     & Decibels relative to full scale             & NGT      & Nominal group technique \\
  DUA      & Data Use Agreement                          & PAD      & Pleasure-Arousal-Dominance \\
  EDA      & Electrodermal activity                      & PPG      & Photoplethysmography \\
  EEG      & Electroencephalography                      & QC       & Quality control \\
  ET       & Eye tracking                                & RMSSD    & Root mean square of successive differences \\
  FAIR     & Findable, Accessible, Interoperable, and Reusable & RR  & R-R (inter-beat) interval \\
  GSR      & Galvanic skin response                      & SAM      & Self-Assessment Manikin \\
  HR       & Heart rate                                  & SCR      & Skin conductance response \\
  HRV      & Heart-rate variability                      & SSE      & Server-Sent Events \\
  IMU      & Inertial measurement unit                   & VAD      & Valence-Arousal-Dominance \\
  KNN      & $k$-nearest-neighbour                       & VFOA     & Visual focus of attention \\
  \addlinespace
  \multicolumn{2}{@{}l}{} & XDF      & Extensible Data Format \\
  \bottomrule
\end{tabular}
\endgroup

\medskip
\noindent\footnotesize\textbf{Note on VAD.}
Throughout this paper \textbf{VAD} denotes \emph{Valence-Arousal-Dominance} (the three dimensions of the affect rating probes).
Voice-activity detection---a distinct signal-processing operation that segments speech from silence in audio recordings---is always written in full to avoid confusion.


\section{Stimuli and Task Orchestration}
\label{app:stimuli}

This appendix describes the complete stimuli suite and session-orchestration
system used during data collection.  All four tasks were delivered by a custom two-tier web application running on the Recording PC: a \emph{display server}
pushed content to (a) five Android tablets (one for each participant and one for the moderator) over local Wi-Fi and (b) a big
screen via HDMI; a \emph{task runner} serialised phase transitions and sent
\ac{LSL} markers for every content change.  \ac{SSE} ensured
near-zero-latency push without polling.

\subsection*{Release Modalities}

\input{tables/modalities}

\subsection*{Participant Demographics}

\begin{figure}[ht]
  \centering
  \includegraphics[width=\linewidth]{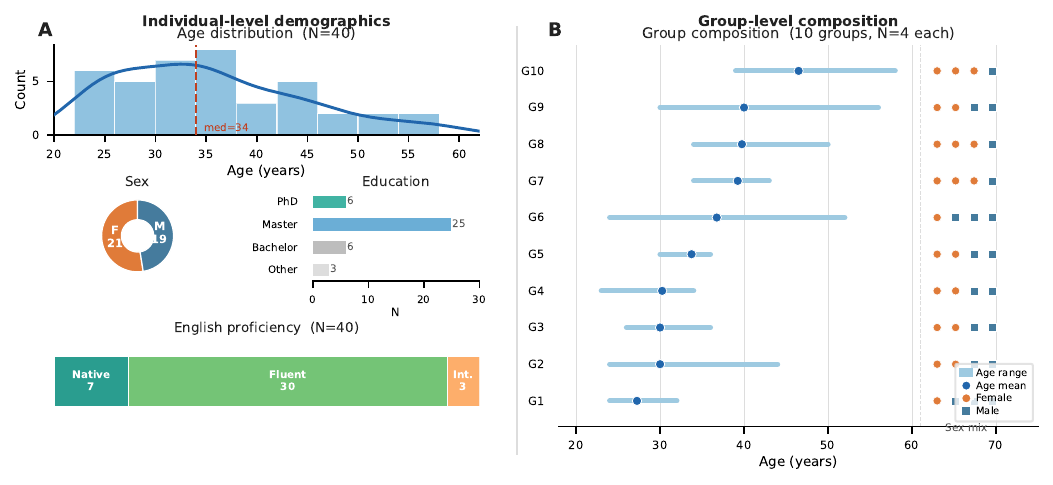}
  \caption{Participant demographics (age, sex, education) and group
    composition across 40 participants in 10 four-person groups.}
  \label{fig:demographics}
\end{figure}

\subsection*{Session Architecture}

A typical session lasted approximately 70 minutes and was preceded by an onboarding phase in which participants were briefed, fitted with the equipment, and calibration procedures were completed. Participants sat around a rectangular table, each with an Android tablet. Content was participant-specific (evidence cards, role cards) and never visible to others. Participants were instructed by the moderator to not show their tablets to others. The display mounted to a wall at the end of a table ("big screen") showed shared task briefs, phases' instructions, and a timer. The moderator controlled the tasks' phases using a web application on a tablet.

\subsection*{Task Descriptions}

\paragraph{T0   Onboarding and Baseline (10 min).}
After providing written informed consent, participants were onboarded by an experimenter who explained the session procedures and sensing equipment, assisted with device setup and calibration, and then conducted a 5‑minute free‑talk warm‑up (self‑introductions; T0). T0 serves as a recording baseline and is excluded from the affective benchmarks.

\paragraph{T1   Hidden-Profile Decision Task (10 min).}
Participants acted as a hiring committee selecting one of three fictional candidates (A, B, C) for an internal ``AI Adoption \& Transformation Specialist'' role. The task followed a hidden‑profile structure: each participant viewed on their tablet a private \emph{evidence card} containing information about all three candidates, with some items identical across all cards (shared evidence) and others present on only one card (unique evidence). Participants were unaware that their information differed from that of other group members, and no single card contained sufficient evidence to identify the normatively correct choice. After 75 seconds of silent individual reading, the group discussed freely. As discussion ended, the moderator instructed the group to choose one person to record the decision on the tablet.

\begin{figure}[ht]
  \centering
  \includegraphics[width=\linewidth]{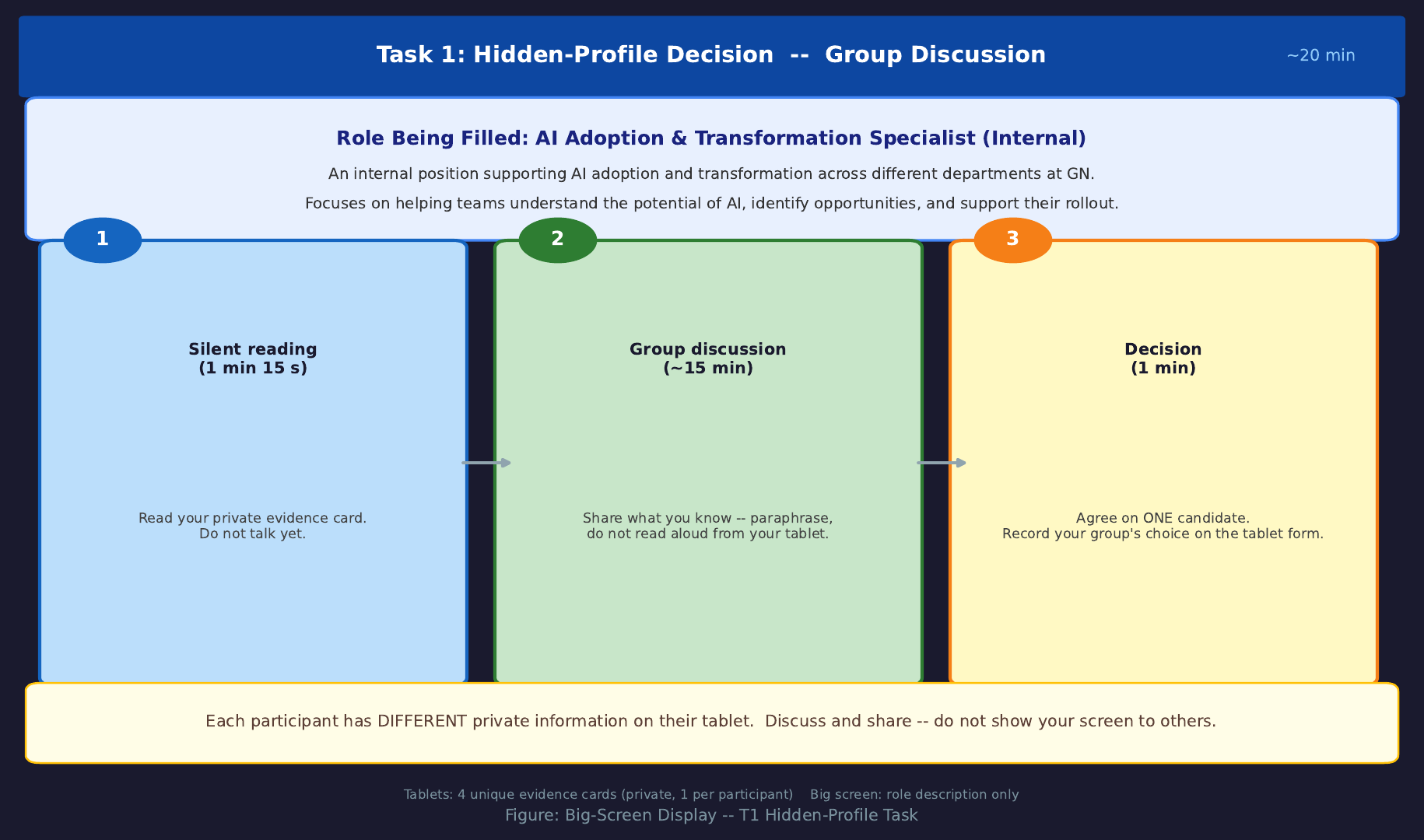}
  \caption{T1 (Hidden-Profile Decision) big-screen layout. The shared display
    shows the task brief, phase instructions, and a countdown timer.
    Participant evidence cards were shown only on individual tablets.}
  \label{fig:stimuli_bigscreen_t1}
\end{figure}

\begin{figure}[ht]
  \centering
  \includegraphics[width=\linewidth]{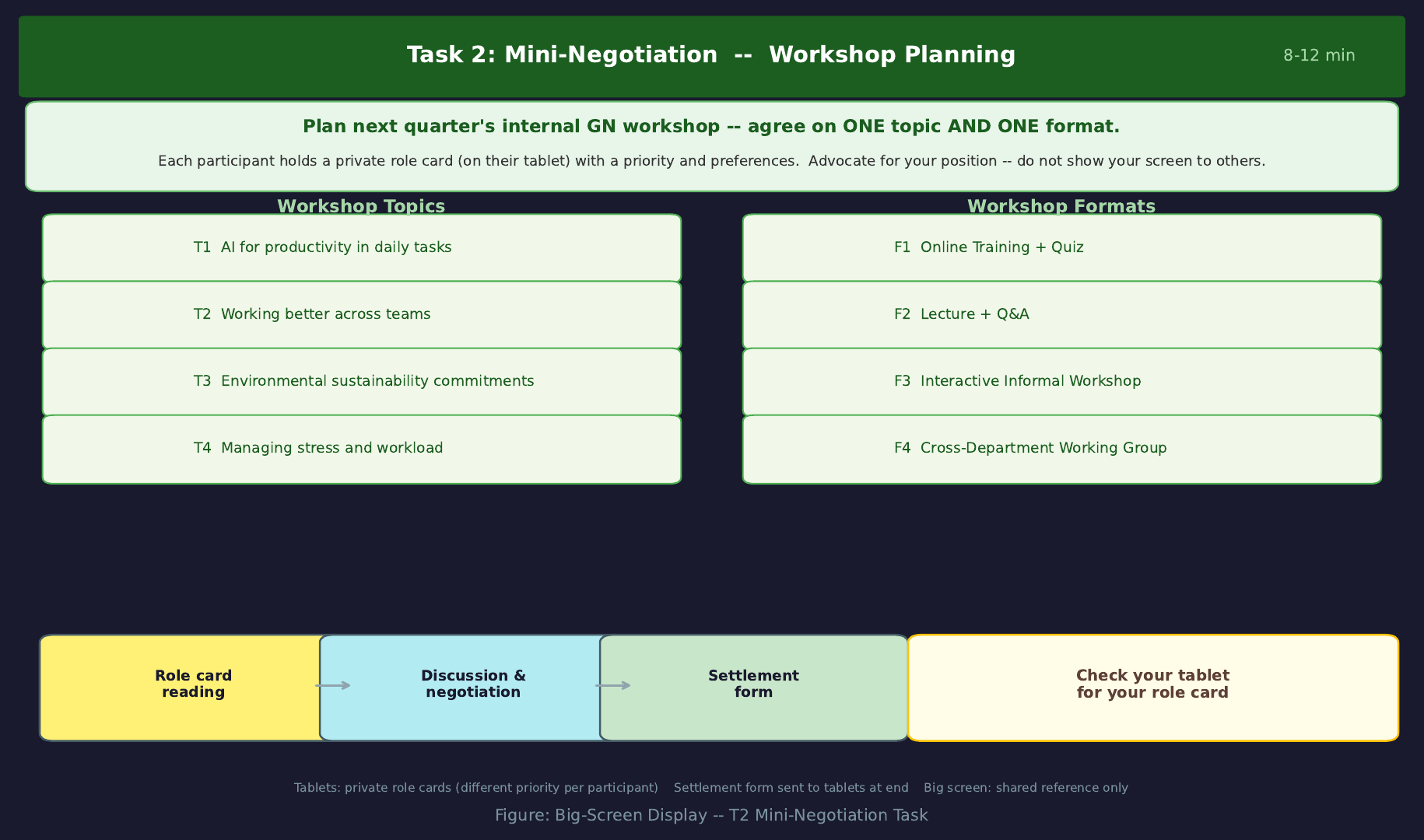}
  \caption{T2 (Mini-Negotiation) big-screen layout. The shared display shows
    the topic and format options; each participant's role card was private.}
  \label{fig:stimuli_bigscreen_t2}
\end{figure}

\begin{figure}[ht]
  \centering
  \includegraphics[width=\linewidth]{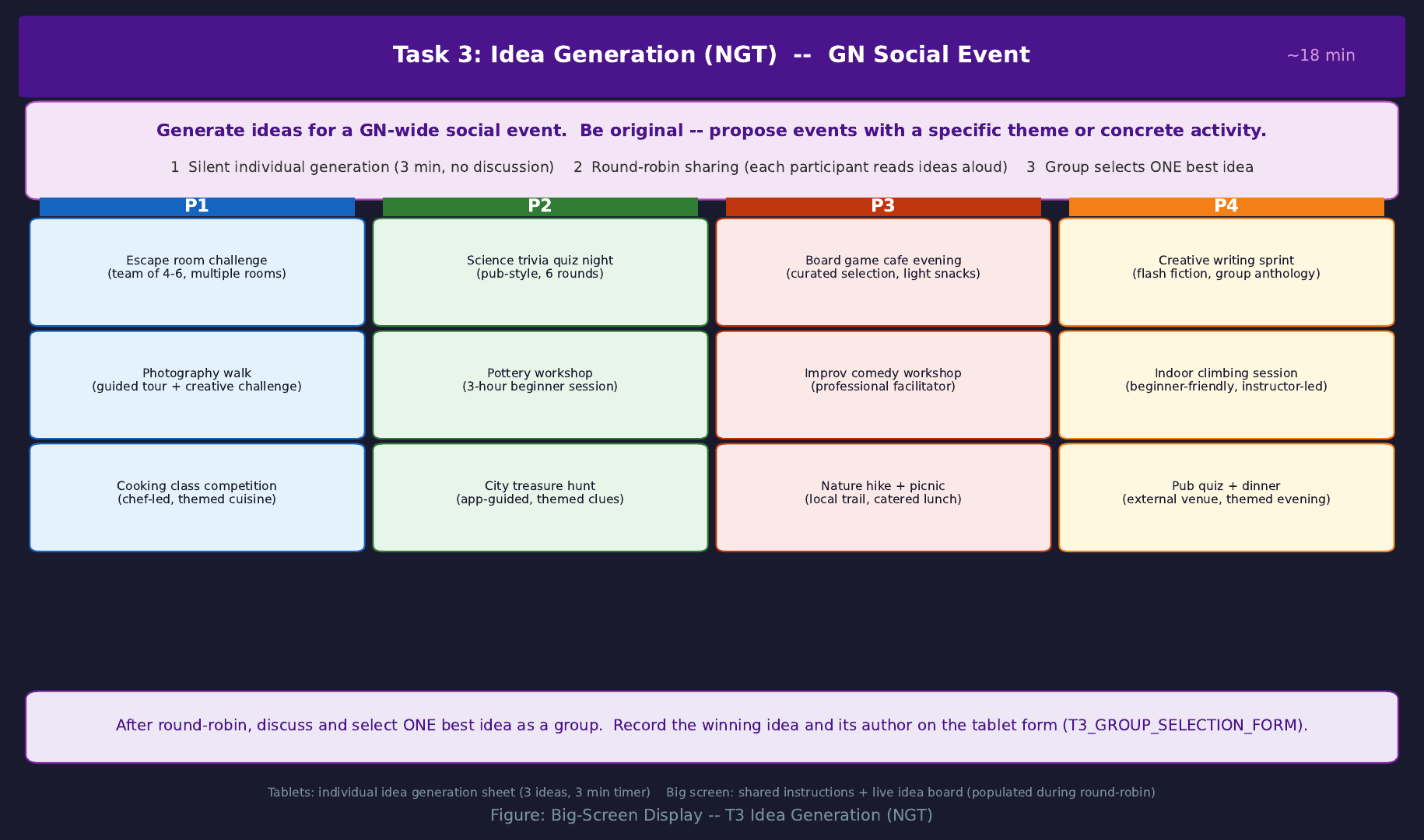}
  \caption{T3 (Idea Generation and Selection) big-screen layout. The shared
    display shows the social-event brief and the group's submitted ideas
    during the discussion phase.}
  \label{fig:stimuli_bigscreen_t3}
\end{figure}

\begin{figure}[ht]
  \centering
  \includegraphics[width=\linewidth]{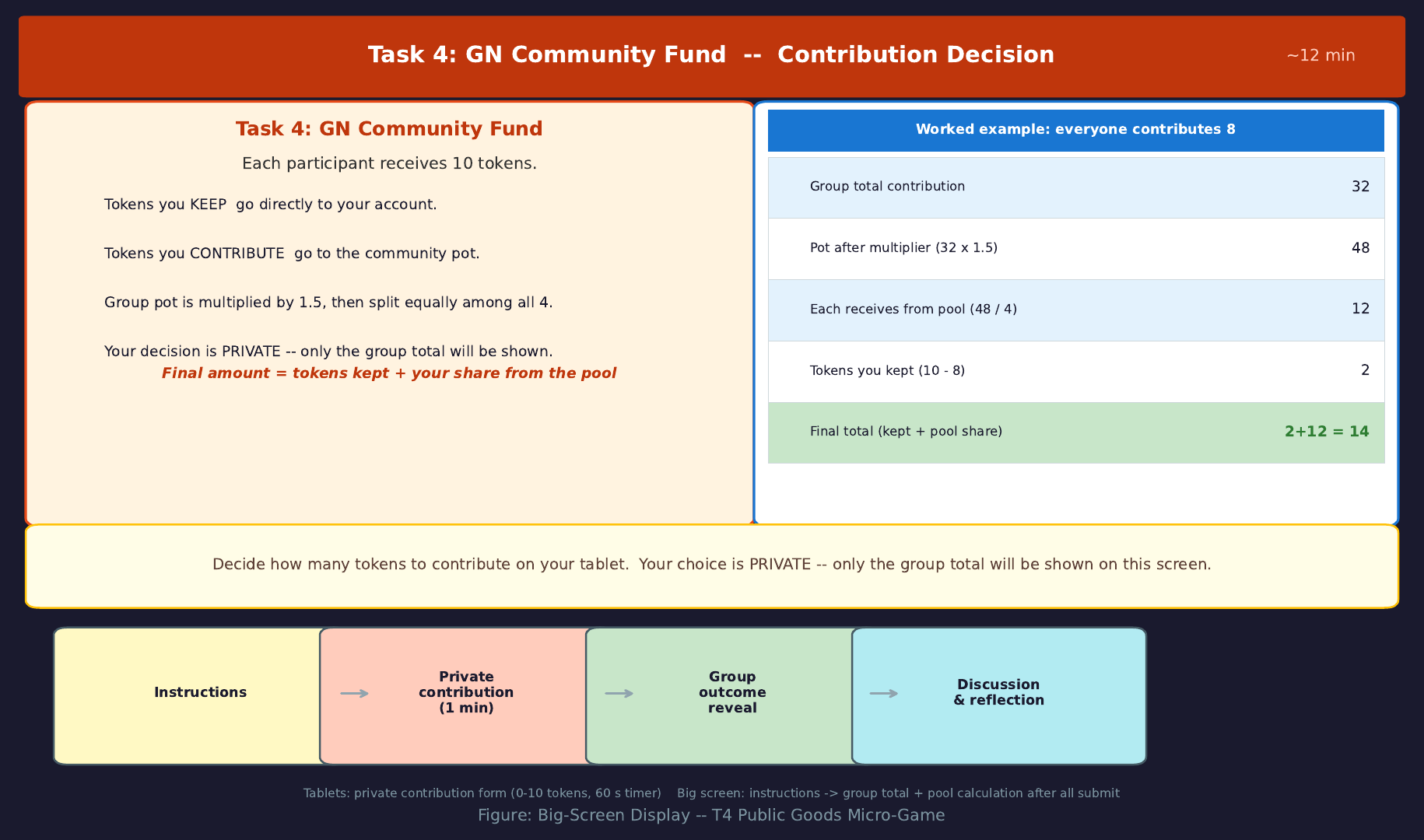}
  \caption{T4 (Public-Goods Micro-Game) big-screen layout. The shared display
    reveals individual contributions after all participants have submitted
    privately, then shows the group payout.}
  \label{fig:stimuli_bigscreen_t4}
\end{figure}

\paragraph{T2   Mini-Negotiation Task (15 min).}
The group jointly planned a quarterly internal workshop, agreeing on both a
\emph{topic} (four options: AI for productivity, cross-team collaboration,
environmental sustainability, stress/workload management) and a \emph{format}
(four options: online training + quiz, lecture + Q\&A, interactive workshop,
cross-department working group).  Each participant received a private
\emph{role card} prescribing a priority dimension (topic or format) and a
preference direction, with an explicit instruction to advocate strongly for
that position. After a 7 minutes discussion, the moderator instructed the group to choose one person to send a settlement form on the tablet. Figures~\ref{fig:stimuli_bigscreen_t1}
through~\ref{fig:stimuli_bigscreen_t4} show the T1--T4 big-screen layouts.

\paragraph{T3   Idea Generation and Selection (18 min).}
Participants generated ideas for a GN-wide social event: first independently writing up to three concrete event ideas on their tablet forms during a 3-minute silent phase, then
discussing all ideas as a group, and finally reaching consensus on one winning idea recorded on a settlement form by one of the participants.
(Figure~\ref{fig:stimuli_bigscreen_t3}).

\paragraph{T4   Public-Goods Micro-Game (12 min).}
Each participant was tasked to imagine they received 10 tokens and could privately decided how many to
contribute to a shared fund with colleagues.  The fund was multiplied by $1.5$ and divided equally among all four participants, creating a classic social-dilemma structure:
\[
  \text{payout}_i = (10 - c_i) + \tfrac{1}{4}\cdot 1.5\sum_j c_j
  \quad c_i \in \{0,\ldots,10\}
\]
Contributions were submitted privately; group outcome was revealed immediately, then discussion followed after individual token contributions were shown on the main display.(Figure~\ref{fig:stimuli_bigscreen_t4}).

\subsubsection*{Task 2 Role Cards}

Each participant received a private role card indicating (i) a primary negotiation priority (topic vs.\ format) and (ii) a secondary preference:

\begin{itemize}
  \item \textbf{Participant 1 (Role Card A).} Primary priority: \textbf{format}. Secondary preference: \textbf{people and culture} topic (Topics~3--4).

  \item \textbf{Participant 2 (Role Card B).} Primary priority: \textbf{topic} (productivity and efficiency; Topics~1--2). Secondary preference: \textbf{independent-learning} format.

  \item \textbf{Participant 3 (Role Card C).} Primary priority: \textbf{format} (independent-learning). Secondary preference: \textbf{productivity and efficiency} topic (Topics~1--2).

  \item \textbf{Participant 4 (Role Card D).} Primary priority: \textbf{topic} (people and culture; Topics~3--4). Secondary preference: \textbf{collaborative and engaging} format.
\end{itemize}

\subsubsection*{Continuous \ac{VAD} Probes}
\label{app:stimuli_vad}

\begin{figure}[ht]
  \centering
  \includegraphics[width=1\linewidth]{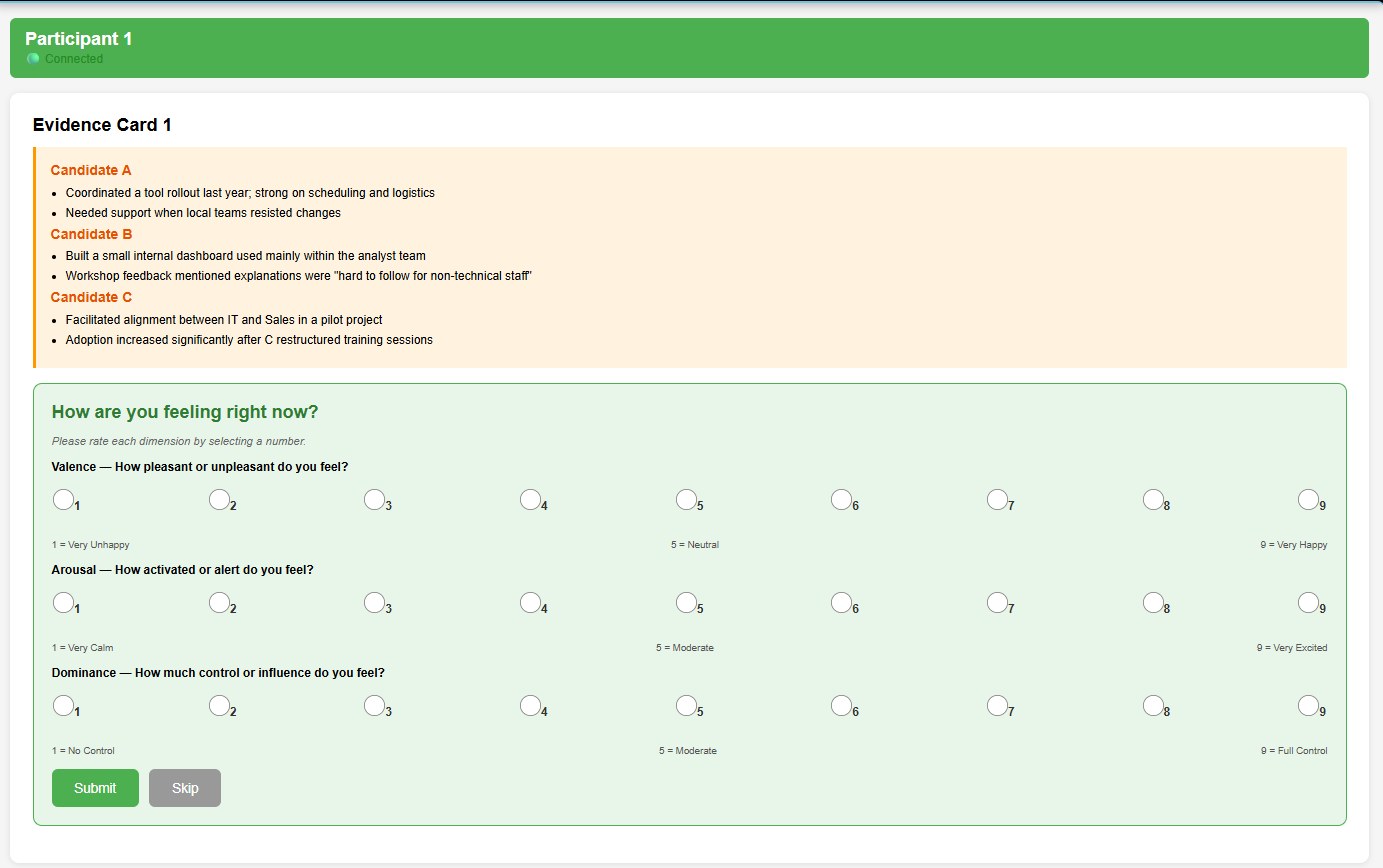}
  \caption{Participant tablet view of a \ac{VAD} affect probe during Task 1.  Valence, Arousal,
    and Dominance are shown simultaneously on a 1--9 Likert scale.}
  \label{fig:stimuli_tablet_vad}
\end{figure}

During each task, participants received periodic affect check-in prompts on
their tablets (Figure~\ref{fig:stimuli_tablet_vad}). Three constructs were
rated on a 1--9 scale:

\begin{itemize}[nosep]
  \item \textbf{Valence:} ``How pleasant or unpleasant do you feel right
        now?'' (1 = very unpleasant, 9 = very pleasant)
  \item \textbf{Arousal:} ``How activated or alert do you feel right now?''
        (1 = very calm / low energy, 9 = very activated / highly alert)
  \item \textbf{Dominance:} ``How much control or influence do you feel right
        now?'' (1 = very little influence, 9 = a great deal of influence)
\end{itemize}

For T4, we omitted the dominance dimension because the anonymous contribution
setup does not provide a clear sense of personal control over others or the
group outcome.

Probes were scheduled at task-specific phase-aligned time points with a small
temporal jitter, so that prompts fell near the midpoints of key interaction
phases (Table~\ref{tab:vad_post_counts}). Across a full four-person
session, this schedule yields up to three \ac{VAD} prompts per participant in
T0--T3 and two prompts in T4.

\subsubsection*{Post-Block Questionnaires}

After each task, participants completed a short individual questionnaire on
their tablets (Table~\ref{tab:posttask_t1} for T1, 
Table~\ref{tab:posttask_t2} for T2, 
Table~\ref{tab:posttask_t3} for T3, and 
Table~\ref{tab:posttask_t4} for T4). These post-block items capture both
self-directed evaluations (e.g., perceived pleasantness, engagement, mental demand, satisfaction, fairness) and other-directed appraisals of each group member.

In the tasks that include trust, each participant rated their trust in each of
the other three group members once, yielding a directed dyadic trust network
with up to $4 \times 3 = 12$ directed trust ratings per task and group.
Similarly, post-task dominance items asked how dominant each of the four
participants (P1--P4) seemed during the task, providing per-seat dominance
ratings from every group member.

\begin{figure}[ht]
  \centering
  \includegraphics[width=1\linewidth]{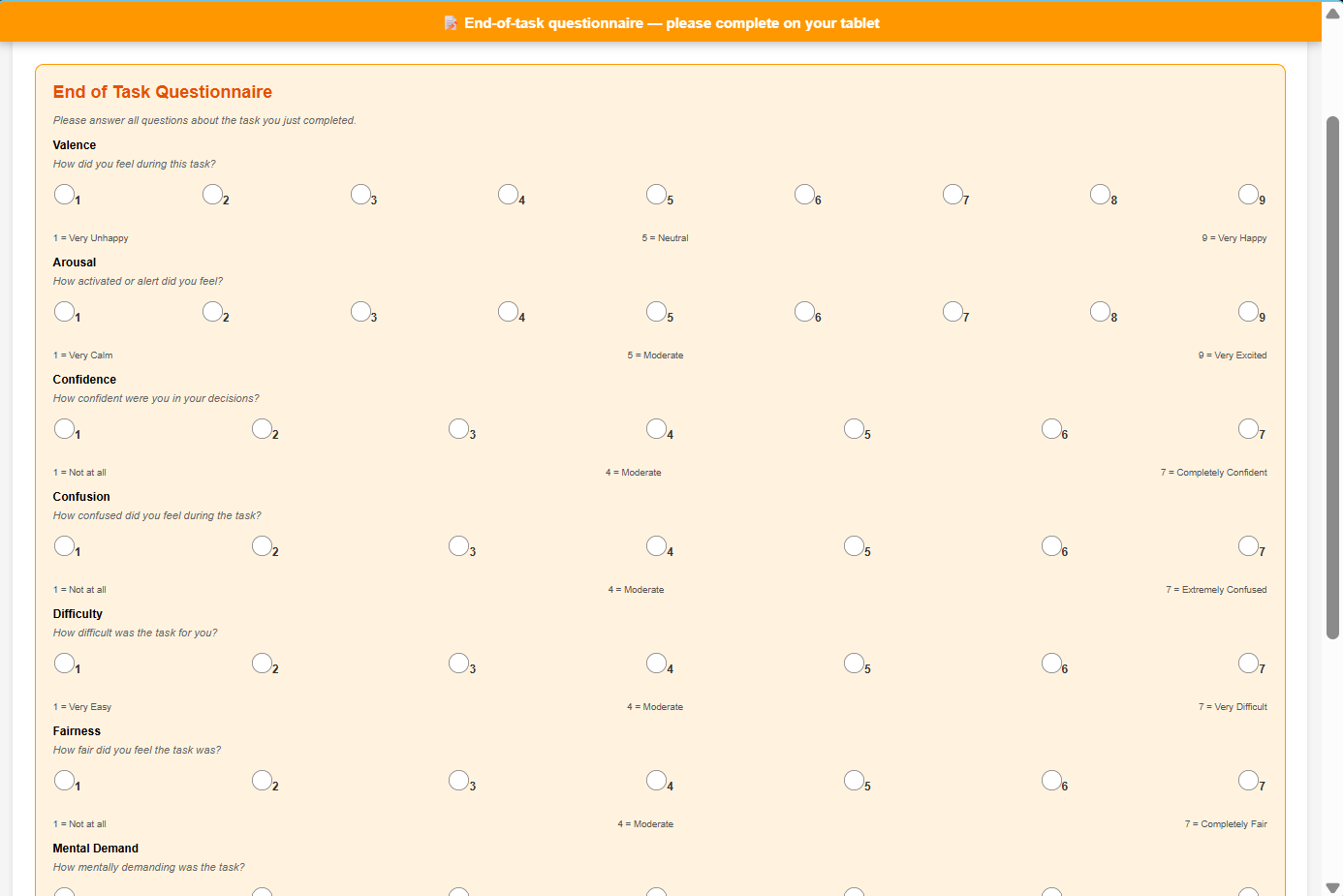}
  \caption{Participant tablet view of the post-block questionnaire (T4 shown, 7 items shown).  All items use a 1--7 Likert scale.
    Participants respond individually.}
  \label{fig:stimuli_tablet_postblock}
\end{figure}

\begin{table}[h]
\centering
\caption{In-task \ac{VAD} probes and post-task items by task.}
\label{tab:vad_post_counts}
\resizebox{\textwidth}{!}{%
\begin{tabular}{l c c c}
\hline
Task &
\ac{VAD} probes (per participant) &
\ac{VAD} dimensions &
\# post-task items \\
\hline
T1: Hidden-profile decision      & 3 & Valence, arousal, dominance          & 18 \\
T2: Mini-negotiation             & 3 & Valence, arousal, dominance          & 15 \\
T3: Idea generation \& selection & 3 & Valence, arousal, dominance          & 16 \\
T4: Public-goods game            & 2 & Valence, arousal (no dominance)      & 13 \\
\hline
\end{tabular}%
}

\smallskip
\raggedright\footnotesize
Note: During onboarding (T0) the moderator sometimes prompted participants to try the \ac{VAD} panel, so a small number of \ac{VAD} submissions appear for T0 in the logs, but there is no formal post-task questionnaire for T0.
\end{table}

\begin{table}[h]
\centering
\caption{Task 1 (Hidden-profile decision) post-task questionnaire items.}
\label{tab:posttask_t1}
\begin{tabular}{p{4cm} p{10cm}}
\hline
\textbf{Label} & \textbf{Question text} \\
\hline
Familiarity -- P1 &
How well did you know P1 before this session? \\
Familiarity -- P2 &
How well did you know P2 before this session? \\
Familiarity -- P3 &
How well did you know P3 before this session? \\
Familiarity -- P4 &
How well did you know P4 before this session? \\
Overall Valence &
Overall, how pleasant or unpleasant did this task feel? \\
Perceived Influence &
How much influence did you feel you had during the task? \\
Engagement &
How engaged/absorbed did you feel during the task? \\
Mental Demand &
How difficult/mentally demanding did the task feel? \\
Team Coordination &
How well did your group coordinate during the task? \\
Voice / Inclusion &
To what extent did you feel that your views were heard and taken into account? \\
Information Sharing &
To what extent do you feel all relevant information was shared within the group before making a decision? \\
Decision Confidence &
How confident are you that your group made the right decision? \\
Equality of Contribution &
How evenly do you think contributions to the discussions were distributed among group members? \\
Dominance -- P1 &
How dominant did Participant 1 seem during the task? \\
Dominance -- P2 &
How dominant did Participant 2 seem during the task? \\
Dominance -- P3 &
How dominant did Participant 3 seem during the task? \\
Dominance -- P4 &
How dominant did Participant 4 seem during the task? \\
Task Check (T1) &
Information that was unique to individual group members influenced the final decision. \\
\hline
\end{tabular}
\end{table}

\begin{table}[h]
\centering
\caption{Task 2 (Mini-negotiation) post-task questionnaire items.}
\label{tab:posttask_t2}
\begin{tabular}{p{4cm} p{10cm}}
\hline
\textbf{Label} & \textbf{Question text} \\
\hline
Overall Valence &
Overall, how pleasant or unpleasant did this task feel? \\
Perceived Influence &
How much influence did you feel you had during the task? \\
Engagement &
How engaged/absorbed did you feel during the task? \\
Mental Demand &
How difficult/mentally demanding did the task feel? \\
Cooperative vs Competitive &
To what extent did the negotiation feel cooperative rather than competitive? \\
Voice / Inclusion &
To what extent did you feel that your views were heard and taken into account? \\
Satisfaction &
How satisfied are you with the outcome? \\
Trust -- Next to You &
How much did you trust the participant next to you? \\
Trust -- In Front &
How much did you trust the participant in front of you? \\
Trust -- At Angle &
How much did you trust the participant sitting at the angle from you? \\
Dominance -- P1 &
How dominant did Participant 1 seem during the task? \\
Dominance -- P2 &
How dominant did Participant 2 seem during the task? \\
Dominance -- P3 &
How dominant did Participant 3 seem during the task? \\
Dominance -- P4 &
How dominant did Participant 4 seem during the task? \\
Task Check (T2) &
The outcome of the negotiation felt mutually beneficial for all parties involved. \\
\hline
\end{tabular}
\end{table}

\begin{table}[h]
\centering
\caption{Task 3 (Idea generation and selection) post-task questionnaire items.}
\label{tab:posttask_t3}
\begin{tabular}{p{4cm} p{10cm}}
\hline
\textbf{Label} & \textbf{Question text} \\
\hline
Overall Valence &
Overall, how pleasant or unpleasant did this task feel? \\
Engagement &
How engaged/absorbed did you feel during the task? \\
Mental Demand &
How difficult/mentally demanding did the task feel? \\
Confidence &
How confident did you feel during the task? \\
Team Coordination &
How well did your group coordinate during the task? \\
Voice / Inclusion &
To what extent did you feel that your views were heard and taken into account? \\
Satisfaction &
How satisfied are you with the outcome? \\
Fairness &
How fair did the outcome feel to you? \\
Psychological Safety &
To what extent did you feel free to propose ideas without worrying about negative reactions? \\
Idea Quality &
To what extent do you think your ideas were at the level of the others? \\
Idea Diversity &
To what extent did the group generate many different and original ideas? \\
Dominance -- P1 &
How dominant did Participant 1 seem during the task? \\
Dominance -- P2 &
How dominant did Participant 2 seem during the task? \\
Dominance -- P3 &
How dominant did Participant 3 seem during the task? \\
Dominance -- P4 &
How dominant did Participant 4 seem during the task? \\
Task Check (T3) &
The group generated many distinct and different ideas during this task. \\
\hline
\end{tabular}
\end{table}

\begin{table}[h]
\centering
\caption{Task 4 (Public-goods game) post-task questionnaire items.}
\label{tab:posttask_t4}
\begin{tabular}{p{4cm} p{10cm}}
\hline
\textbf{Label} & \textbf{Question text} \\
\hline
Overall Valence &
Overall, how pleasant or unpleasant did this task feel? \\
Social Evaluation Concern &
To what extent were you concerned that the other participants would evaluate or judge your choice? \\
Fairness &
How fair did the outcome feel to you? \\
Trust -- Next to You &
How much did you trust the participant next to you? \\
Trust -- In Front &
How much did you trust the participant in front of you? \\
Trust -- At Angle &
How much did you trust the participant sitting at the angle from you? \\
Expectation Match &
To what extent did the outcome match your expectations about how much people would contribute? \\
Regret &
To what extent do you regret the contribution you chose to make? \\
Dominance -- P1 &
How dominant did Participant 1 seem during the task? \\
Dominance -- P2 &
How dominant did Participant 2 seem during the task? \\
Dominance -- P3 &
How dominant did Participant 3 seem during the task? \\
Dominance -- P4 &
How dominant did Participant 4 seem during the task? \\
Task Check (T4) &
When making my choice, I was concerned about how the other participants would evaluate or judge me. \\
\hline
\end{tabular}
\end{table}

Hard behavioural outcomes are stored in the \texttt{beh/} BIDS subfolder: the T1 candidate selected (A/B/C; C is the normatively correct hidden-profile choice), the T2 agreed topic--format pair, the T3 winning idea with authorship attribution, and each participant's T4 token contribution (0--10).

\section{Extended Limitations and Caveats}
\label{app:extended_limitations}

\paragraph{Physiology precision.}
PPG-derived HRV at $\approx$25~Hz imposes a $\approx$40~ms RMSSD quantisation floor;
wearable HRV is not equivalent to ECG-grade measurement (\Cref{app:ppe_detail}).

\paragraph{Pupil and audio comparability.}
Pupil summaries remain sensitive to illumination and display context; the released
audio features are not baseline-normalised because T0 free-talk audio is insufficiently
controlled, so multimodal comparisons mix relative physiology/pupil features with
absolute audio features.

\paragraph{Survey ceiling and role coverage.}
Some social-evaluative ratings are high and stable across tasks (e.g.\ voice/inclusion
means 5.85$\pm$1.23 in T1, 4.77$\pm$1.66 in T2, 5.90$\pm$1.11 in T3), which could
reflect genuine inclusivity or acquiescence bias.
The dataset does not contain direct measures of emergent leadership, expertise
attribution, or role crystallisation.

\paragraph{Voice re-identification.}
Close-talk audio carries voice re-identification risk, motivating the separate
access tier and DUA described in \Cref{sec:ethics}.

\paragraph{Egocentric gaze.}
Released gaze is in each participant's own scene-camera frame; cross-participant
gaze-target analysis requires room-frame alignment not included in this release.

\paragraph{Deferred modalities.}
Multi-camera video, marker-assisted 3D pose, and room-frame gaze alignment
were recorded or piloted but are reserved for a future release.

\paragraph{No clinical ground truth.}
The dataset must not be used for diagnosis, mental-health inference,
personnel evaluation, or surveillance-style deployment.

\section{BFI-44 Scoring and Item List}
\label{app:bfi44}

BFI-44 domain scores are computed as item means after applying the standard
reverse-coding rules.
The release datasheet includes the item-level mapping, reverse-coded item
list, and missing-response policy.

\section{Audio T0 Baseline Reliability}
\label{app:audio_t0_baseline}

Audio features are used in two normalization regimes depending on benchmark type.
Within-person benchmarks (B0--B3: task classification, affective state) apply
within-person z-score normalisation across tasks T1--T4 only, capturing
task-induced variance.
Between-person benchmarks (B4--B5: personality, contribution) use absolute
audio features (no baseline subtraction) to preserve between-person amplitude
differences required for trait prediction. When we tested normalising
audio features by T0 free-talk baseline across all participants, every tested
feature showed increased variance compared to absolute form, indicating T0 introduction of noise.

\paragraph{Why T0 is unreliable.}
The unstructured T0 session differs fundamentally from structured tasks T1--T4 (Table~\ref{tab:audio_t0_reliability}). Absence of task constraints leads to natural heterogeneity: some participants dominate, others remain quiet; vocal energy and pitch vary with conversational mood; social dynamics are unpredictable. T0 is not a neutral biological baseline but rather reflects initial social calibration, novelty effects, and potential social anxiety---orthogonal to individual vocal traits. Third, audio metrics (speaking fraction, pause count, overlap) are group properties, not individual traits; T0 free-talk creates role instability (variable who-speaks-vs-listens patterns) that cannot be attributed to individual differences.

\paragraph{Implementation consequence.}
Because T0 normalisation universally increases variance, audio for B4--B5
(personality, between-person) is retained in absolute form, preserving
between-subject amplitude variation.
Separately, T1--T4 within-person normalisation is retained for B0--B3 because
those four structured tasks share similar role clarity and behavioral demands.
This reflects principled design: physiological features (cardiac, electrodermal)
remain trait-stable across tasks and benefit from T0 baseline correction, while
interaction-structure metrics require comparable task context.

\begin{table}[t]
  \centering\small
  \caption{Audio feature variance comparison: absolute form vs.\ T0-normalised
    form across all 40 participants and 200 participant-task rows.
    All features show increased standard deviation when normalised by T0
    baseline, confirming that T0 free-talk is insufficiently controlled to serve
    as a reliable per-participant reference. Consequently, audio is kept in
    absolute form for between-person benchmarks (B4--B5).}
  \label{tab:audio_t0_reliability}
  \begin{tabular}{@{}lrrrp{0.35\linewidth}@{}}
    \toprule
    \textbf{Audio Feature} & \textbf{Abs.\ SD} & \textbf{Delta-T0 SD} &
      \textbf{Ratio} & \textbf{Interpretation} \\
    \midrule
    speaking\_fraction & 0.036 & 0.354 & 9.8$\times$ &
      T0 baseline highly variable; signals person was quiet then talkative,
      not stable trait \\
    pause\_count & 24.821 & 214.685 & 8.6$\times$ &
      Pausing naturally dependent on uncontrolled T0 social dynamics \\
    overlap\_fraction & 0.030 & 0.078 & 2.6$\times$ &
      Group turn-taking in unstructured T0 creates confounding \\
    speech\_rate\_proxy & 0.680 & 1.106 & 1.6$\times$ &
      Vocal tempo variance inflated; T0 normalisation adds noise \\
    \bottomrule
  \end{tabular}
\end{table}

\section{Synchronisation Pipeline Detail}
\label{app:sync}

Task windows are derived from experiment-control markers written to the
session event spine.
Per-modality feature tables are then joined to these windows using timestamp
filters and participant/seat mappings.
The release preserves enough timing provenance to audit alignment at multiple
levels: event-marker windows, per-device frame logs, progress streams, and the
derived \texttt{sync\_metadata.json} sidecars. In the current pipeline,
frame-log start-spread analyses target sub-10~ms AV alignment, while DPA audio
requires explicit correction for a measured hardware-clock drift of about
0.04~ms/s relative to the XDF/LSL clock. The audio splitter therefore fits a
per-microphone linear time map instead of relying on a single median offset,
yielding practical task-level residual error below 5~ms in documented
sessions.

\section{Preprocessing Steps}
\label{app:preprocessing}

The five-step pipeline applied for all benchmarks (\Cref{sec:benchmarks})
is described below; \Cref{tab:preprocessing} summarises the quantitative
impact of each step across the 136 active participant-task rows.

\subsection*{Pipeline Steps}

\input{tables/preprocessing_steps}

\paragraph{\ac{ET} quality gating.}
Eye-tracking rows where $>$50\% of samples are missing have pupil features
set to NaN; rows where the gaze-valid fraction is $<$50\% have gaze features
set to NaN.

\paragraph{Physiological plausibility gating.}
Values outside credible windows are replaced with NaN
(\ac{HR} $\notin$[40,180]~bpm; RMSSD $\notin$[10,300]~ms;
EDA $\notin$[0,25]~$\mu$S; pupil $\notin$[1.5,9.0]~mm;
pitch $\notin$[5,55]~semitones; delta features have tighter windows).
In total 67 values are gated across all modalities.

\paragraph{Winsorisation.}
After plausibility gating, values are clipped at $\pm3\,\sigma$ per feature.
Only 32 values are clipped, confirming plausibility bounds removed the most
extreme outliers.

\paragraph{Within-person robust z-score.}
For state and task benchmarks, features are centred by the participant's
median and scaled by $1.4826\times\text{MAD}$ across T1--T4 rows.
This is the most impactful step: pupil dilation Cohen's~$d$ for T2 vs.\ T1
rises from 0.28 to 1.13; speaking-fraction $d$ doubles from 0.44 to 0.88.
This transform is computed before leave-one-group-out splitting, so held-out
participants contribute their own unsupervised T1--T4 distribution statistics
to test-time normalisation. We retain it because the paper's benchmark goal is
dataset characterisation rather than online deployment, but it should be read
as a mild leakage source for B0--B3d. For between-person targets (B4a--B5),
within-person z-scoring is not applied.

\paragraph{KNN imputation.}
A $k=5$ nearest-neighbour imputer is fit inside each \ac{LOGOCV} training fold
and applied to the held-out group, preventing missing-data leakage.

\subsection*{Feature Selection}

Feature selection was applied globally to the full 136-row active-task dataset
(T1--T4) before any \ac{LOGOCV} split using two criteria applied sequentially:
(1) features with more than 50\% missing values were dropped;
(2) among remaining features, highly correlated pairs ($|r|>0.95$) were
reduced by greedy removal, retaining the first feature in each correlated
cluster.
After the full pipeline, 35 features survive selection.
Five composite biomarker features (\texttt{biomarker\_*}) are excluded from
the benchmarks: they were computed as internal linear combinations of
physiological and pupil features already in the retained set
(e.g.\ \texttt{biomarker\_cognitive\_load}\,=\,mean(z\_pupil,
z\_eda, $-$z\_rmssd)), and their presence alongside raw source components
makes benchmark attribution ambiguous.

\subsection*{Retained Feature Set}

\Cref{tab:retained_features} lists all 35 retained features grouped by
modality. The four annotation process-metadata features (bottom group) are
listed for completeness but are \emph{excluded} from the 31-feature benchmark
set; see Leakage Sources below.

\begin{table}[h]
  \centering
  \caption{Retained feature set after global feature selection (35 features total;
    31 used in all reported benchmarks). ``$\Delta$T0'' = value relative to the
    T0 free-talk baseline. The annotation group is excluded from benchmark runs.}
  \label{tab:retained_features}
  \small
  \setlength{\tabcolsep}{4pt}
  \begin{tabular}{@{}p{0.44\linewidth}p{0.49\linewidth}@{}}
    \toprule
    \textbf{Feature} & \textbf{Description} \\
    \midrule
    \multicolumn{2}{@{}l}{\textit{Physiology (8 features)}} \\
    \quad\texttt{hr\_mean\_bpm\_delta\_t0}         & Mean \ac{HR} relative to T0 \\
    \quad\texttt{hr\_sd\_bpm}                       & \ac{HR} variability (SD) \\
    \quad\texttt{hrv\_rmssd\_ms}                    & \ac{HRV} RMSSD from PPG \\
    \quad\texttt{hrv\_quality\_score}               & PPG signal quality index \\
    \quad\texttt{eda\_tonic\_mean\_delta\_t0}       & Tonic EDA $\Delta$T0 \\
    \quad\texttt{eda\_phasic\_rate\_hz\_delta\_t0}  & Phasic EDA event rate $\Delta$T0 \\
    \quad\texttt{eda\_phasic\_mean\_delta\_t0}      & Phasic EDA amplitude $\Delta$T0 \\
    \quad\texttt{eda\_scr\_count}                   & \ac{SCR} count \\
    \addlinespace
    \multicolumn{2}{@{}l}{\textit{Motion / Temperature (3 features)}} \\
    \quad\texttt{temp\_mean\_delta\_t0}             & Skin temperature $\Delta$T0 \\
    \quad\texttt{accel\_motion\_mean}               & Mean wrist-acceleration magnitude \\
    \quad\texttt{motion\_high\_fraction}            & Fraction of high-motion samples \\
    \addlinespace
    \multicolumn{2}{@{}l}{\textit{Eye-tracking (5 features)}} \\
    \quad\texttt{pupil\_left\_mean}                 & Left pupil diameter (mm) \\
    \quad\texttt{pupil\_right\_mean}                & Right pupil diameter (mm) \\
    \quad\texttt{pupil\_std}                        & Pupil diameter SD \\
    \quad\texttt{pupil\_slope\_per\_s}              & Linear pupil dilation slope \\
    \quad\texttt{pupil\_mean\_delta\_t0}            & Mean pupil diameter $\Delta$T0 \\
    \addlinespace
    \multicolumn{2}{@{}l}{\textit{Audio (15 features)}} \\
    \quad\texttt{audio\_energy\_mean\_x}            & Mean frame energy \\
    \quad\texttt{audio\_energy\_sd\_x}              & Frame energy variability \\
    \quad\texttt{audio\_hnr\_mean\_x}               & Harmonics-to-noise ratio \\
    \quad\texttt{audio\_jitter\_mean\_x}            & Pitch period jitter \\
    \quad\texttt{audio\_mean\_unvoiced\_segment\_s} & Mean unvoiced segment duration \\
    \quad\texttt{audio\_mean\_voiced\_segment\_s\_x}& Mean voiced segment duration \\
    \quad\texttt{audio\_overlap\_fraction\_x}       & Overlapping-speech fraction \\
    \quad\texttt{audio\_pause\_count}               & Pause event count \\
    \quad\texttt{audio\_pitch\_mean\_x}             & Mean fundamental frequency \\
    \quad\texttt{audio\_pitch\_sd\_x}               & Pitch variability \\
    \quad\texttt{audio\_shimmer\_mean\_x}           & Amplitude shimmer \\
    \quad\texttt{audio\_speaking\_fraction\_x}      & Speaking-time fraction \\
    \quad\texttt{audio\_speaking\_time\_s}          & Total speaking time (s) \\
    \quad\texttt{audio\_speech\_rate\_proxy\_x}     & Syllable-rate proxy \\
    \quad\texttt{audio\_voiced\_segments\_per\_sec\_x} & Voiced-segment density \\
    \midrule
    \multicolumn{2}{@{}l}{\textit{Annotation process-metadata (4 features, excluded from benchmarks)}} \\
    \quad\texttt{answers\_n}                        & Responses submitted \\
    \quad\texttt{ann\_total\_events\_n}             & Total annotation events \\
    \quad\texttt{ann\_response\_postblock\_n}       & Post-block form submissions \\
    \quad\texttt{ann\_event\_span\_s}               & Annotation event time span (s) \\
    \bottomrule
  \end{tabular}
\end{table}

\subsection*{Leakage Sources}

Three deviations from a fully fold-internal pipeline are acknowledged:

\begin{enumerate}
  \item \textbf{Within-person z-score (affects B0--B3d).}
    Normalisation statistics (median, \ac{MAD}) are computed over all four task
    rows per participant before the \ac{LOGOCV} split, so held-out participants
    contribute their own unsupervised T1--T4 distribution to test-time
    normalisation. No directional label bias, but within-person effect sizes
    are inflated. Not applied for between-person targets (B4a--B5).

  \item \textbf{Global feature selection.}
    Missing-rate and correlation-based filtering is applied to the full dataset
    before splitting. Cannot favour any particular label direction; impact on
    \ac{AUC} is negligible, but represents a strict-protocol deviation.

  \item \textbf{Annotation process-metadata features.}
    The four annotation features (\texttt{answers\_n},
    \texttt{ann\_total\_events\_n},
    \texttt{ann\_response\_postblock\_n}, \texttt{ann\_event\_span\_s})
    are excluded from the 31-feature benchmark set.
    \texttt{ann\_event\_span\_s} and \texttt{answers\_n} vary systematically
    by task (T2 has more \ac{VAD} probes and a longer annotation span),
    creating a direct shortcut for the B0 task-classification sanity check.
    \texttt{ann\_response\_postblock\_n} is a data-completeness proxy
    structurally correlated with annotation-derived B3/B\_trust targets.
    B0 accuracy rises to 0.734 when these features are included
    (vs.\ reported 0.641), confirming the exclusion is conservative.
    All feature-importance figures (\Cref{app:ablation}) and benchmark
    scripts exclude these four features.
\end{enumerate}

\subsection*{Technical Caveats}

\phantomsection\label{app:ppe_detail}
\paragraph{PPG-derived HRV precision.}
\texttt{hrv\_rmssd\_ms} is derived from PPG peak detection at $\approx$25~Hz.
At this sampling rate, the minimum detectable \ac{RR} interval difference is
$\approx$40~ms, setting a quantisation floor on RMSSD.
Spearman correlations between \texttt{hrv\_rmssd\_ms} and post-block
self-report labels are weak ($r=-0.09$ with engagement, $r=0.21$ with
arousal, both $p>0.05$; $n=53$--$67$), consistent with the limited
precision of wrist PPG-derived HRV at this resolution.
The feature is retained because it provides the only autonomic vagal-tone
proxy available in the release; users requiring ECG-grade HRV should treat
\texttt{hrv\_rmssd\_ms} as an approximate indicator rather than a precise
measurement.

\paragraph{EDA motion contamination.}
The mean \texttt{motion\_high\_fraction} across all participant-task rows is
0.0015 (SD\,0.0018), indicating that elevated wrist-motion periods represent
a negligible fraction of recorded time in this cohort.
The feature was dropped during global feature selection due to near-zero
variance, confirming that motion-elevated EDA periods are rare enough to
have minimal influence on the benchmark features.

\section{Extended Dataset Characterization}
\label{app:worked_extended}

\input{tables/autonomic_worked_example}

\subsection*{Personality and multimodal response}
Spearman correlations between the five BFI traits and four participant-level
behavioural means (valence, arousal, engagement, mental demand) yield 20
tested pairs ($n=40$). After \ac{BHFDR} correction, no association survives at
$q<0.05$. The nominally strongest associations before correction are
Agreeableness with arousal ($r=0.37$, $p=0.017$) and Agreeableness with
engagement ($r=0.35$, $p=0.028$). The previously reported Openness--speech-rate
correlation ($r=-0.40$, $n=31$) remains present but does not survive the full
correction.
These participant-level correlations treat individuals as independent observations; a linear mixed-effects model with random intercepts per participant and task would account for the nested structure and yield calibrated confidence intervals the recommended approach for future work with a larger cohort.
The complete correlation table is in the release datasheet;
BFI results are released primarily as covariates for future work given that
$n=40$ is severely underpowered for personality--behaviour associations.
Per-group BFI-44 trait profiles are available in the release metadata.

\paragraph{Language proficiency}
BFI-44 was administered in English.
English proficiency among the 40 participants is: native speaker (10, 25\%),
fluent non-native (24, 60\%), intermediate (6, 15\%).
BFI-44 has not been validated specifically for intermediate second-language
respondents, and item interpretation may vary with proficiency.
The six intermediate-proficiency participants represent a small subgroup;
their BFI responses are included in the release but should be interpreted
with this caveat.
English proficiency ($\texttt{english\_proficiency}$) is released as a
covariate to support sensitivity analyses; at $n=40$ this dataset is
underpowered to detect proficiency-moderated effects reliably.

\subsection*{Familiarity}
Participants rated their prior acquaintance with each other on a 7-point Likert scale (1 = ``never met'', 
7 = ``very familiar'') before the session began. 
The distribution was heavily skewed toward low familiarity: 
70.0\% of responses fell below 4, and 62.4\% were exactly 1 (``never met''). 
Overall mean familiarity was $M=2.79$ (SD\,2.56), indicating that participants were 
predominantly strangers, which is consistent with the recruitment strategy of assigning 
groups from volunteers across different departments within the organization.
This low baseline familiarity setting reduces confounds from pre-existing team dynamics.

\subsection*{Task outcomes and interaction dynamics}
\begin{figure}[h]
    \centering
    \includegraphics[width=0.95\textwidth]{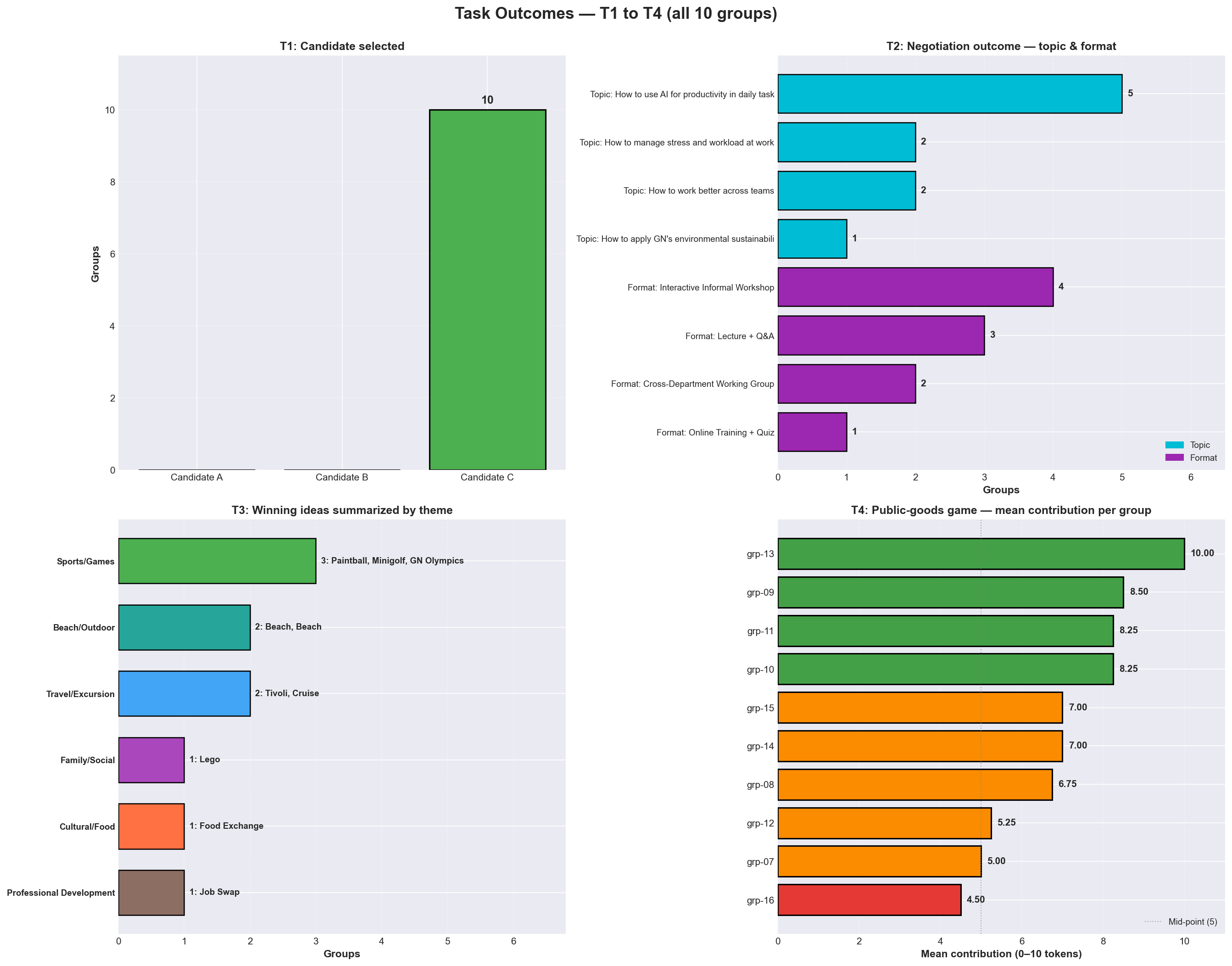}
    \caption{
        \textbf{Task Outcomes Dashboard (T1--T4).}
        Panel 1: Hidden-profile decision outcome (T1). 
        All 10 groups selected Candidate C. 
        Panel 2: Mini-negotiation outcome (T2). 
        Topics and formats are colour-differentiated. 
        Most common: ``AI for productivity'' (5 groups).
        Panel 3: Idea Generation outcome (T3). 
        Winning ideas are grouped by themes.  
        Panel 4: Public-Goods Contribution (T4). 
        Per-group mean contributions (0--10 scale) sorted by value. 
        Colour intensity indicates cooperation level (red $<5$, orange 5--7.5, green $\geq 7.5$). 
        Overall mean 7.05 (SD 1.76) reflects strong cooperative tendency.
    }
    \label{fig:task_outcomes}
\end{figure}

\begin{figure}[h]
    \centering
    \includegraphics[width=0.95\textwidth]{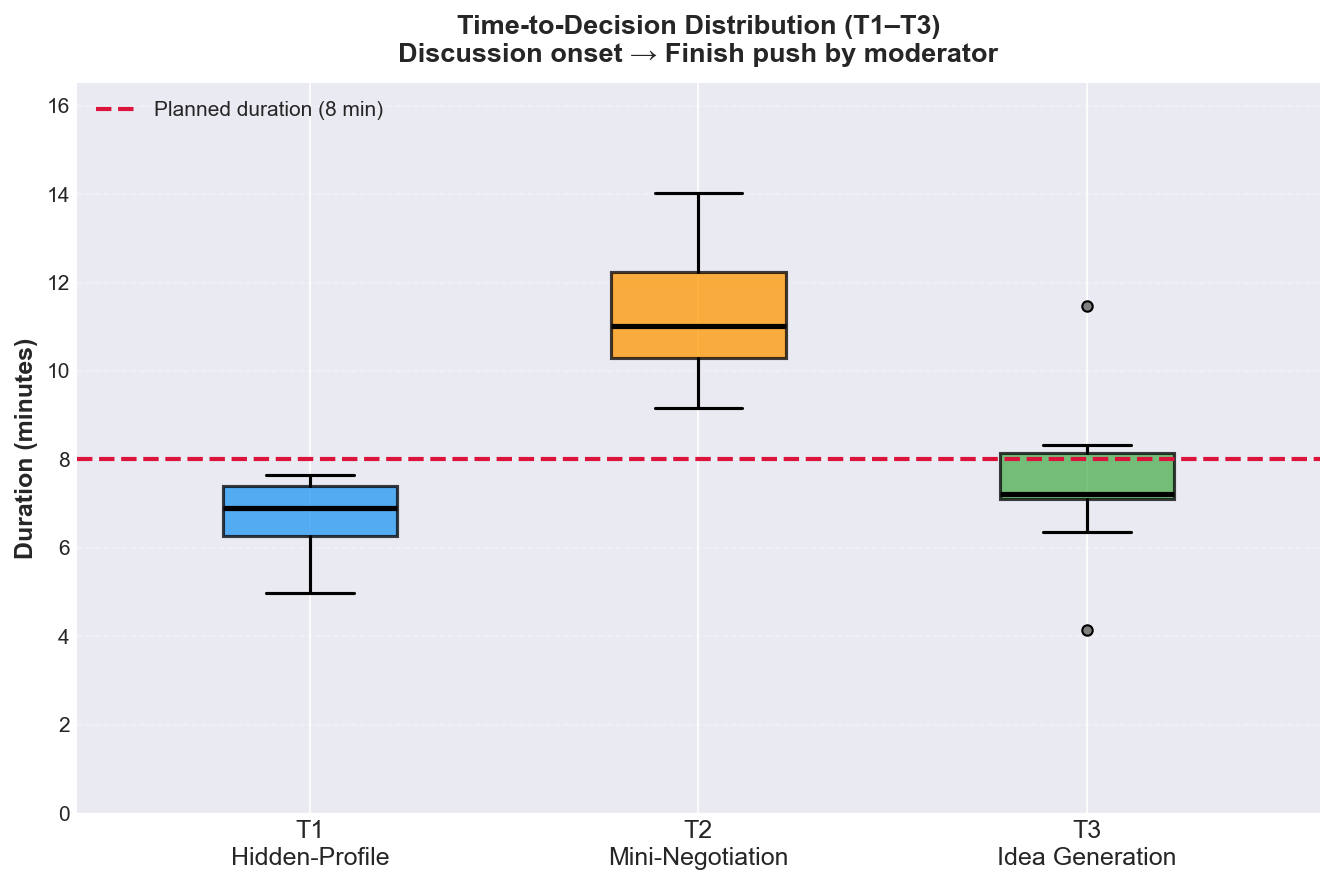}
    \caption{
        \textbf{Time-to-decision durations (T1--T3).}
        Boxplots show discussion durations from onset to the moderator ``finish'' prompt for T1--T3.
        The dashed horizontal line marks the nominal 8-minute guideline; groups typically overran this
        duration, especially in T2. T4 is excluded because its discussion phase did not terminate in a
        single group decision. Durations were derived directly from \texttt{events\_grp-XX.tsv} files
        (one per group) by filtering moderator \texttt{push\_content} events for the relevant task and
        computing the difference between the \texttt{role\_card} start-phase marker and the
        corresponding \texttt{finish} marker. Onsets are machine-recorded LSL timestamps (seconds
        relative to session start), so these durations reflect exact event-log timing rather than
        manual annotation.
    }
    \label{fig:timing_decision}
\end{figure}

The four structured tasks produce quantifiable group outcomes that can be linked to multimodal
signals. \cref{fig:task_outcomes} summarizes the per-group results across all decision tasks.
\begin{itemize}
    \item \textbf{T1:}
    All 10 groups reached consensus on \textit{Candidate C}, the correct choice. One group (grp-07) submitted multiple response entries for the same participant, reflecting participants' misunderstanding of instructions.
    
    \item \textbf{T2:}
    All 10 groups ($n=10/10$) reached settlement on both topic and format.
    \emph{Discussion timing:} 
    Despite the 8-minute guideline, all 10 groups overran, with mean discussion duration 
    11.4\,minutes (SD\,1.6, range 9.2--14.1\,min).
    Although specific timings for each phase in each task were fixed, the moderator was instructed not to interrupt the organic conversation flow and to prompt rapid closure as soon as possible after timer expiration. As a result, groups often overran the nominal phase timings (see \cref{fig:timing_decision}).
    
    \item \textbf{T3 (Idea Generation via Nominal Group Technique):}
    All 10 groups generated ideas and voted on a winner.
    Each group's final selection is logged as \texttt{idea\_author} 
    (format: ``P\textit{X} --- Idea \textit{N}''), which links back to the participant's submitted 
    \texttt{idea\_N} text fields. 
    
    \item \textbf{T4 (Public-Goods Micro-Game):}
    Individual contributions (tokens contributed out of 10-token endowment) yield 
    $M=7.05$ (SD\,1.76, range 4.50--10.00 per group mean).
    High overall contribution level (70.5\% of endowment) indicates strong cooperative tendency.
    Per-group means vary from 4.50 (grp-16, conservative) to 10.00 (grp-13, fully cooperative),
    providing continuous behavioural targets for cooperation-prediction benchmarks.
\end{itemize}

\subsection*{Self-report dynamics}
In-task self-reports of valence and arousal, measured on a 9-point range, reveal distinct affective signatures across the four tasks.
\emph{Valence} shows a clear task effect: Mean ratings were highest in T0 (baseline/introductions; 
$M=7.59$), T3 (idea generation; $M=7.36$), and T4 (micro-game; $M=7.22$), but dropped significantly 
in T1 (hidden-profile decision; $M=6.87$) and most notably in T2 (negotiation; $M=5.76$).
This pattern aligns with task demands: T2 requires adversarial negotiation and format selection, 
which introduces conflict and uncertainty, thereby dampening positive affect.
In contrast, \emph{arousal} shows an opposite pattern: Baseline arousal in T0 was lowest ($M=5.22$), 
with all task phases elevating arousal similarly (T1: $M=6.12$, T2: $M=6.35$, T3: $M=6.21$, T4: $M=6.13$).
This indicates that structured group tasks themselves activate heightened engagement, irrespective of affective valence. The asymmetry between valence and arousal suggests that conflict (T2) dampens pleasure while maintaining engagement.

\subsection*{Post-task survey profiles}
\Cref{tab:construct_task_coverage} maps each post-block survey construct
to the tasks in which it was administered.
Core items (engagement, mental demand, overall valence, per-seat dominance
rating) appear in all four active tasks and support full cross-task
comparisons.
Trust items appear only in T2 and T4; satisfaction only in T2 and T3; voice
and inclusion in T1--T3 but not T4.
Researchers planning cross-task analyses of a specific construct should
verify availability in this table first.

\input{tables/construct_task_coverage}

\emph{Mental demand} differed significantly across tasks. T2 (negotiation) elicited the highest mental load 
($M=4.79$, SD\,1.42), compared to T1 ($M=3.20$, SD\,1.68) and T3 ($M=3.65$, SD\,1.84). 
This aligns with T2's dual demand: participants must simultaneously negotiate content (topic and format) 
and social dynamics (consensus-seeking under time pressure).

\emph{Satisfaction} ratings reveal recovery after difficult negotiation: T2 satisfaction 
($M=5.09$, SD\,1.57) was significantly lower than T3 ($M=6.09$, SD\,1.02), suggesting that the 
post-negotiation idea-generation task provided a more positive experience, possibly due to reduced 
interpersonal conflict and the creative freedom of idea generation.

\emph{Voice inclusion} measures whether the participant felt like their views were being heard in the dicussion. It rated consistently high in T1 ($M=5.80$, SD\,1.23) and T3 ($M=5.84$, SD\,1.22), but declined in T2 ($M=4.81$, SD\,1.64), the negotiation task. This suggests that asymmetric participation or perceived dominance emerges during negotiation, consistent with the higher mental demand and lower satisfaction ratings for that task.

\emph{Trust} was captured as \emph{seat-directed interpersonal ratings}: each participant rated the three other group members (front, next, angle) on trust-related items (7-point scale). For example, a participant in position P1 rates trust toward P4 (front), P2 (next), and P3 (angle). We first computed each participant's trust score as the mean of their three seat-directed ratings, then compared T2 and T4.
At the participant level ($n=40$ paired observations), trust averaged $M=4.98$ (SD\,1.09) after T2 and 
$M=5.25$ (SD\,1.24) after T4, with mean change $\Delta=+0.27$ (SD\,1.64). The paired test was not 
significant, $t(39)=1.03$, $p=.31$. At the group level ($n=10$; mean across participants within group), 
all groups were above midpoint in both tasks (10/10 in T2; 10/10 in T4), and 5/10 groups showed a net 
increase from T2 to T4. The paired group-level change was also non-significant, $t(9)=1.51$, $p=.165$. 
The directional increase is consistent with a cooperative rebound from negotiation (T2) to collective-action play (T4), but inferential results are not reliable at current sample size and should be interpreted as 
exploratory.

\subsection*{Conversation dynamics and turn-taking}
The per-seat close-talk layout enables voice-activity-derived
conversation-structure summaries.
\Cref{tab:audio_turn_taking_summary} reports transcript-derived turn-taking
statistics by task.
T0 produces the most turns ($\bar{n}=54.8$) but the shortest mean duration
(2.19~s).
T1 and T2 show the longest individual turns (4.41 and 4.19~s) and the lowest
overlap fractions (0.125 and 0.107), consistent with deliberative or
competitive conversational floors.
T3 stands out for long response gaps (5.19~s) consistent with the nominal
group technique.
These patterns motivate benchmarks B9--B11 (\Cref{app:extended_benchmarks}).

\input{tables/audio_turn_taking_summary}

\begin{figure}[ht]
  \centering
  \includegraphics[width=\linewidth]{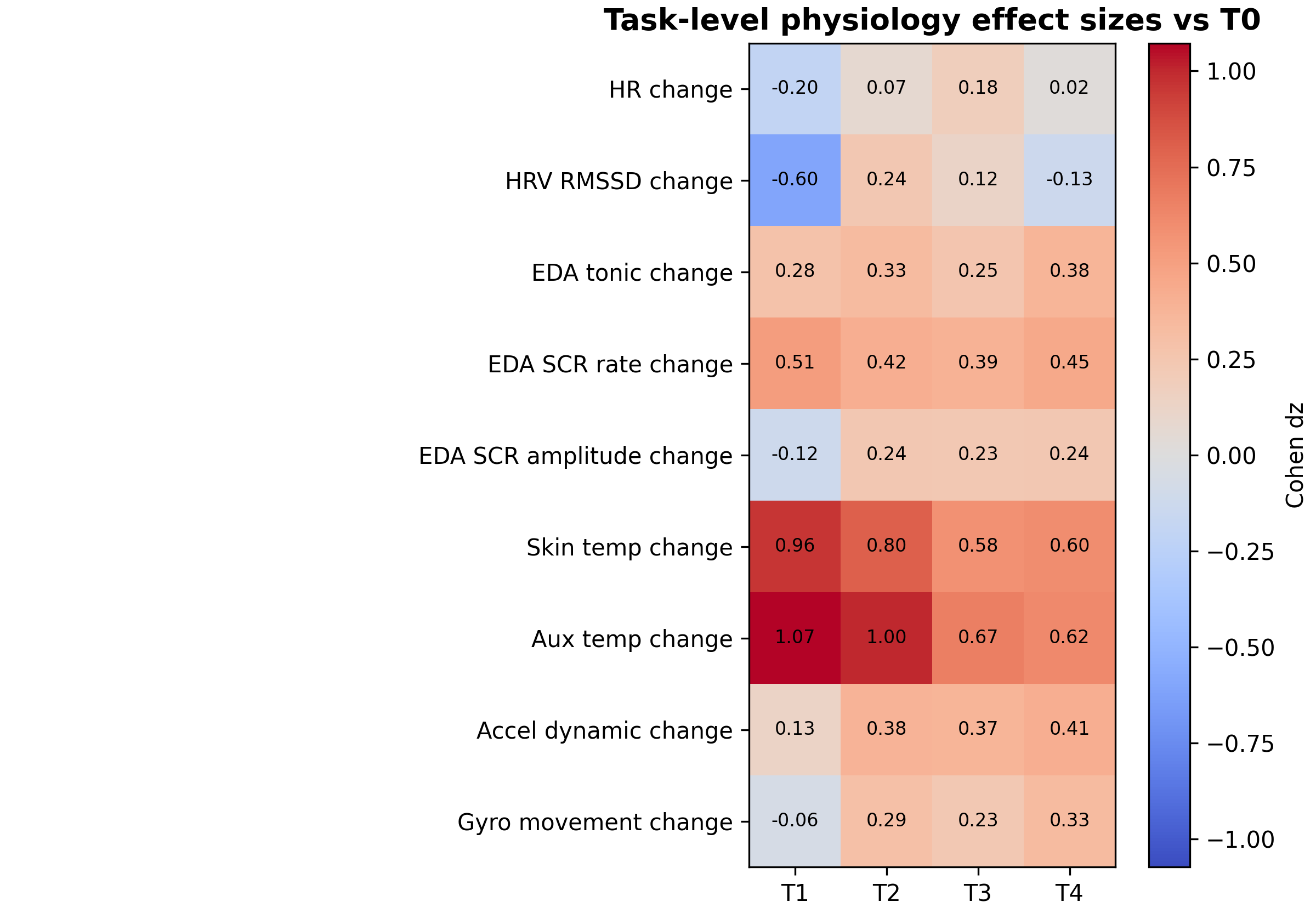}
  \caption{Task-level physiological feature effect sizes (Cohen's~$d$)
    across modalities. T2 shows sustained EDA elevation; T3 shows high \ac{HR}
    and \ac{SCR}; T4 shows moderate increases.}
  \label{fig:physio_effects}
\end{figure}

\begin{figure}[ht]
  \centering
  \includegraphics[width=\linewidth]{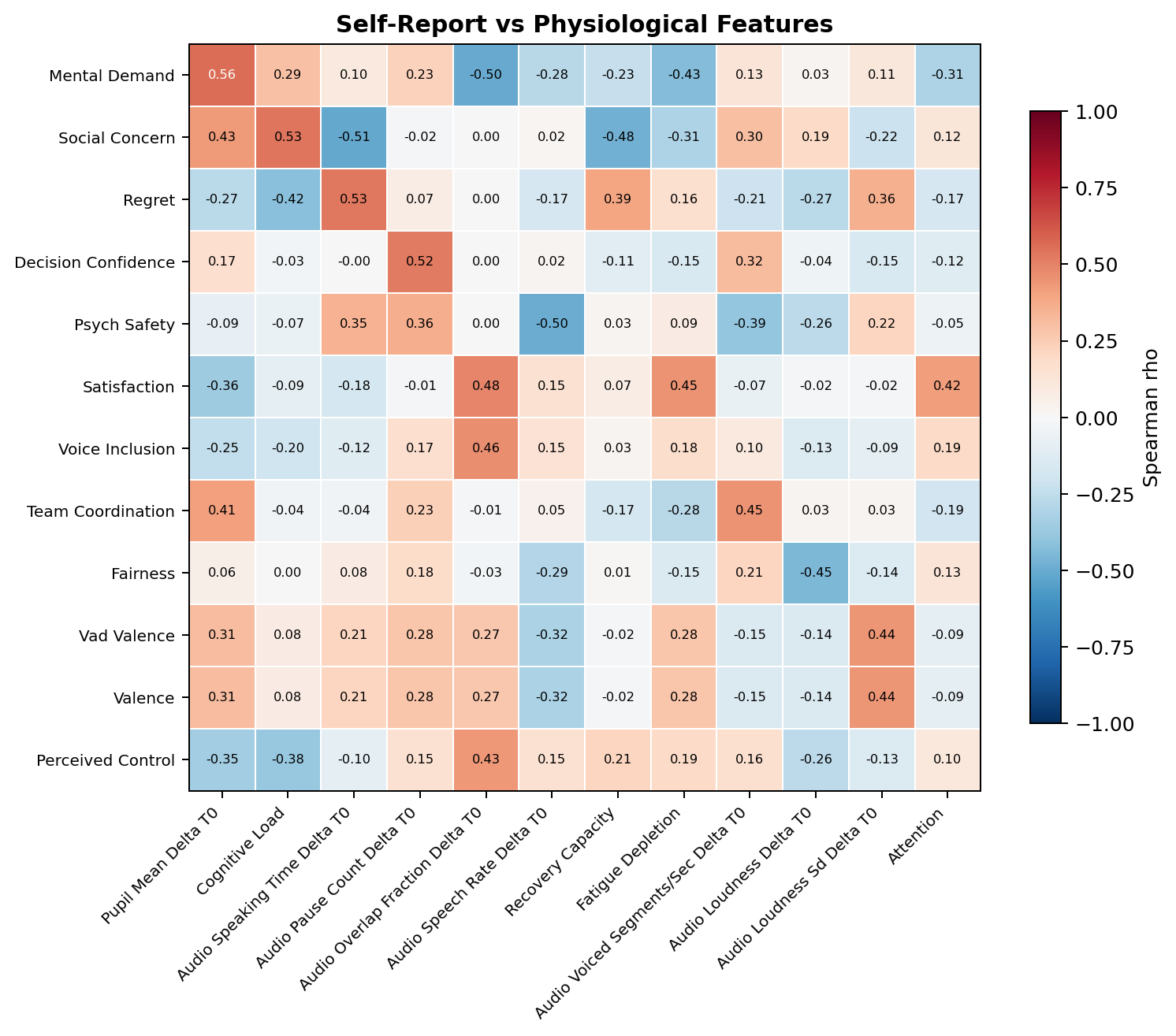}
  \caption{Cross-modal Spearman correlation matrix (physiological and audio
    features vs.\ self-report annotation targets) on 136 active
    participant-task rows with \ac{BHFDR} correction.
    \texttt{audio\_overlap\_fraction} ($r=-0.62$) and pupil dilation are the
    dominant predictors of mental demand, motivating benchmark B3a.}
  \label{fig:cross_modal}
\end{figure}

\section{Extended Benchmarks: Sequential Conversation Tasks}
\label{app:extended_benchmarks}

The per-seat close-talk recordings support three additional benchmarks
requiring per-turn event extraction from voice-activity-detection segmentation.

\emph{B9: Next-speaker prediction.}
Given a context window ending at a turn boundary, predict which of the
remaining three participants takes the floor next (four-class; majority
baseline $\approx0.33$).
Inputs would include last-$N$-second physiology windows, gaze fixation
vectors, and prosodic tail features.

\emph{B10: Turn-taking point detection.}
Given a rolling 2-second window of per-seat voice-activity-detection output, predict whether
a turn transition will occur within the next 500~ms (binary; AUC~+~F1).

\emph{B11: Overlap onset prediction.}
Given the current speaker's prosodic trajectory, predict whether the next
transition involves simultaneous speech (binary; positive fraction
$\approx 0.15$ based on \Cref{tab:audio_turn_taking_summary}).

All raw data for B9--B11 is in place; per-turn voice-activity-detection event extraction is
deferred to a future release.


\section{Benchmark Interpretation Notes}
\label{app:bench_notes}

\begin{figure}[ht]
  \centering
  \includegraphics[width=\linewidth]{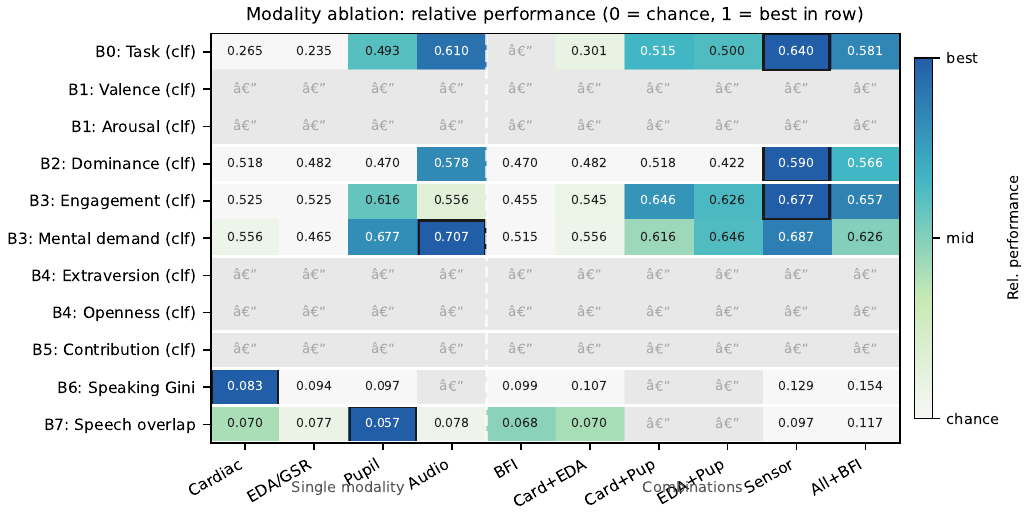}
  \caption{Modality ablation heatmap. Colour encodes performance relative to
    chance (white = chance, dark teal = best).
    Audio dominates task classification; audio and pupil are jointly strongest
    for cognitive-state targets.
    See \Cref{tab:ablation} for numerical details.}
  \label{fig:ablation_heatmap}
\end{figure}

Per-benchmark commentary for the results in \Cref{tab:benchmark_results};
modality ablation (\Cref{fig:ablation_heatmap}) and feature importance
(\Cref{fig:feat_imp_ranked}) are also presented here.

\paragraph{B0: Task-label classification (sanity check).}
Accuracy 0.641 (SD\,0.132, 95\%~CI [0.55, 0.73]) vs.\ 0.265 baseline ($n=136$).
\emph{The previously published figure of 0.734 was inflated} by annotation
process-metadata features (\texttt{ann\_event\_span\_s}, \texttt{answers\_n})
that vary systematically by task; see \Cref{app:preprocessing}.
B0 confirms the tasks are behaviourally distinguishable under LOGO-CV,
not that audio is especially strong for affect inference.

\paragraph{B1a/B1b: Valence and arousal.}
Valence AUC~0.657; arousal AUC~0.528 ($n=107$).
Near-chance arousal reflects a post-block temporal mismatch: probes capture
retrospective state rather than the peak physiological response.
An event-contingent approach at decision moments would better align labels with signals.

\paragraph{B2: Dominance.}
AUC~0.499 ($n=83$, 95\%~CI [0.37, 0.62]) near chance.

\paragraph{B3a/B3b: Mental demand and engagement (strongest Level-1 result).}
B3a AUC~0.719 ($n=99$); B3b AUC~0.591 ($n=99$).
A single \texttt{audio\_overlap\_fraction\_x} feature achieves AUC~0.766 for mental
demand cognitive-load detection is driven by conversational-floor dynamics,
not physiology alone.
B3a and B3b load on different feature profiles (\Cref{fig:feat_imp_ranked}):
B3a is dominated by audio overlap and HR; B3b by pupil slope and pitch.
This within-dataset dissociation confirms demand and engagement are not proxies for one construct.

\paragraph{B3c: Satisfaction (new, $\dagger$).}
AUC~0.571 ($n=60$, T2 and T3 only).
T3 satisfaction is significantly higher than T2 ($t=3.33$, $p=0.001$), but
the demand--satisfaction dissociation weakens B3c relative to B3a with annotation process-metadata features excluded.

\paragraph{B3d: Trust (new, $\dagger$).}
AUC~0.562 pooled ($n=60$); AUC~0.679 T4-only ($n=28$).
Near-zero within-group variance in T2 trust (SD\,=\,0.185) makes the T2 fold-local
split near-random; the cooperative T4 context recovers the signal.

\paragraph{B4a/B4b/B4c: Personality traits (challenge).}
All three are near or below chance under LOGO-CV; test folds of four participants
make AUC estimates inherently unstable regardless of true signal.
Spearman correlations on the full sample confirm signal presence:
\texttt{pupil\_right\_mean} $\times$ Agreeableness $r=+0.51$ ($p=0.008$).
A two-feature model restricted to T2 achieves AUC~0.625 for Agreeableness.
The B4 trio defines a well-motivated but presently unsolvable challenge.

\paragraph{B5: T4 contribution.}
AUC~0.429 ($n=28$, SD~0.290); highly imbalanced fold structure.
The continuous raw contribution (mean 7.05/10, SD~2.88) is the preferred
future modelling target over binary split.

\paragraph{B6a/B6b: Speaking Gini.}
B6a MAE~0.089 vs.\ baseline 0.086 (group-mean features).
B6b Ridge MAE~0.102 vs.\ baseline 0.088 (SD features) both below naive.
A binary classifier on raw speaking-fraction SD alone achieves AUC~0.952
(95\%~CI [0.857, 1.000]), confirming the signal is definitively present;
the Ridge failure is overfitting at $n=28$ rows.

\paragraph{B7: Speech-overlap fraction.}
MAE~0.063 vs.\ baseline 0.060; same mean/variance mismatch as B6a.

\paragraph{Modality contributions.}
\Cref{fig:ablation_heatmap} reports LOGO-CV performance under ten feature-subset conditions.
Audio dominates task-structure and cognitive-demand detection.
Pupil adds independent signal for cognitive state beyond audio alone.
Physiology contributes primarily to arousal-sensitive targets (B1a, B2).

\paragraph{Feature-level importance.}
\Cref{fig:feat_imp_ranked} shows top-15 features by mean normalised
$|\text{coefficient}|$ across LOGO-CV folds.
\texttt{audio\_overlap\_fraction\_x} ranks \#1 in five benchmarks.
B3a and B3b load on entirely different profiles (audio vs.\ pupil+pitch),
confirming the demand--engagement dissociation.
B3d trust is led by speaking time and shimmer rather than floor competition.

\begin{figure*}[ht]
  \centering
  \includegraphics[width=\linewidth]{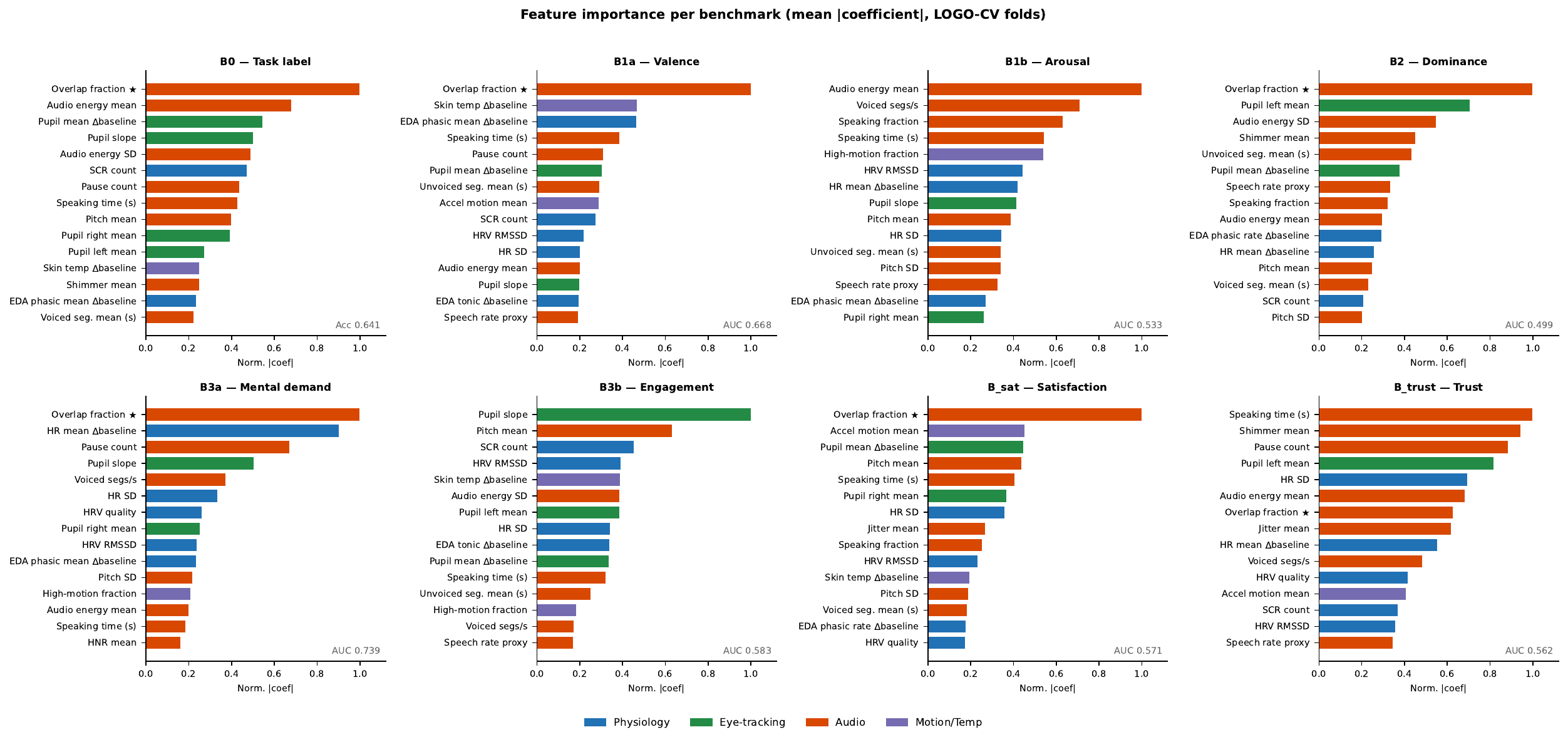}
  \caption{Per-benchmark ranked feature importance: top-15 features by mean
    normalised $|\text{coefficient}|$ across LOGO-CV folds (31-feature set).
    Bar colour: \textcolor[HTML]{2171b5}{\textbf{Physiology}},
    \textcolor[HTML]{238b45}{\textbf{Eye-tracking}},
    \textcolor[HTML]{d94801}{\textbf{Audio}},
    \textcolor[HTML]{756bb1}{\textbf{Motion/Temp}}.
    \textbf{$\star$}~=~\texttt{audio\_overlap\_fraction\_x}.}
  \label{fig:feat_imp_ranked}
\end{figure*}
\section{Full Ablation Table}
\label{app:ablation}

\input{tables/ablation_table}

\begin{figure}[ht]
  \centering
  \includegraphics[width=\linewidth]{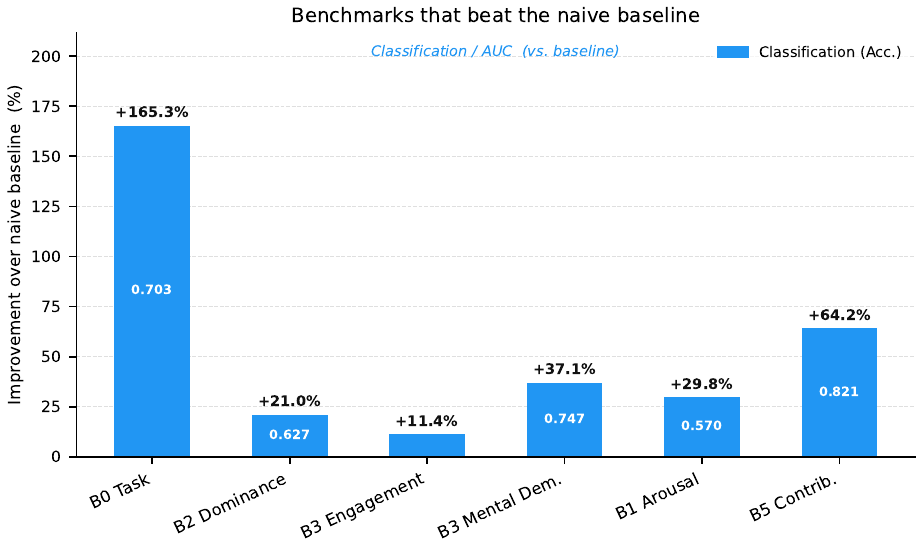}
  \caption{Feasibility baselines for the 31-feature set (biomarker composites
    and annotation process-metadata features excluded; leakage-corrected).
    Each bar shows percentage improvement over the majority-class
    baseline or naive regressor; benchmarks at or below baseline are omitted.
    B3a mental demand is the clearest above-chance participant-state target;
    B1a valence is moderate; B2 dominance and B5 are near or below chance in the
    clean analysis.
    B4a--B4c (personality) are systematically \emph{below} chance and are not shown.
    B6a--B7 (group-level) remain at or below naive MAE and are not shown.
    See \Cref{tab:benchmark_results} for full results with CIs.}
  \label{fig:benchmark_overview}
\end{figure}


\begin{figure}[ht]
  \centering
  \includegraphics[width=\linewidth]{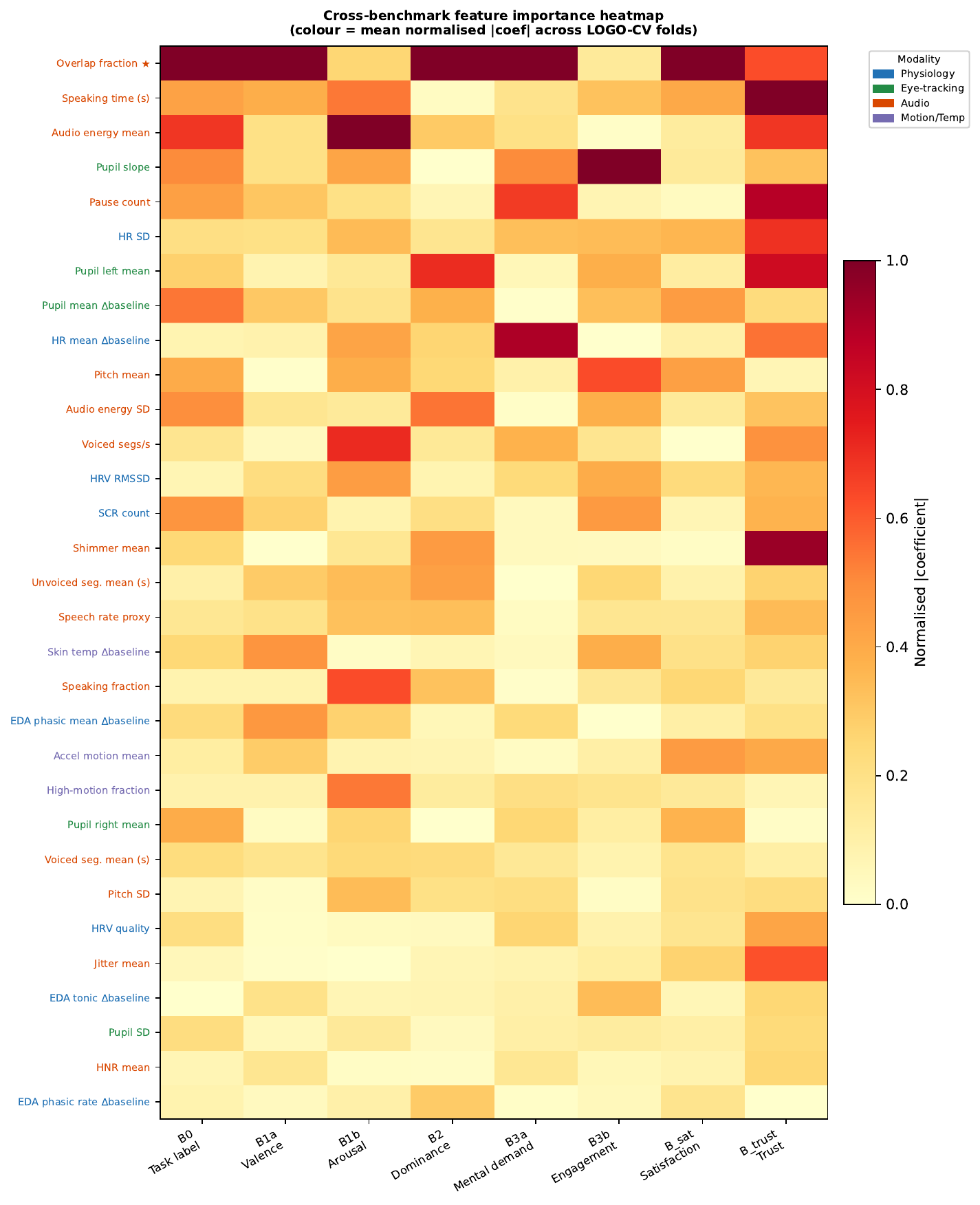}
  \caption{Cross-benchmark feature importance heatmap.
    Rows are all 31 sensor/behavioural features ranked by mean importance across benchmarks
    (annotation process-metadata features excluded; see text);
    columns are benchmark targets.
    Cell colour encodes normalised $|\text{coef}|$ (dark~=~high importance);
    y-axis label colour indicates modality group.
    Audio dominates the left benchmarks (B0, B3a); physiology and pupil
    features differentiate the affective state benchmarks (B1a, B2).
    Generated by \texttt{paper/analysis/feature\_importance.py}.}
  \label{fig:feat_imp_heatmap}
\end{figure}

\begin{figure}[ht]
  \centering
  \includegraphics[width=\linewidth]{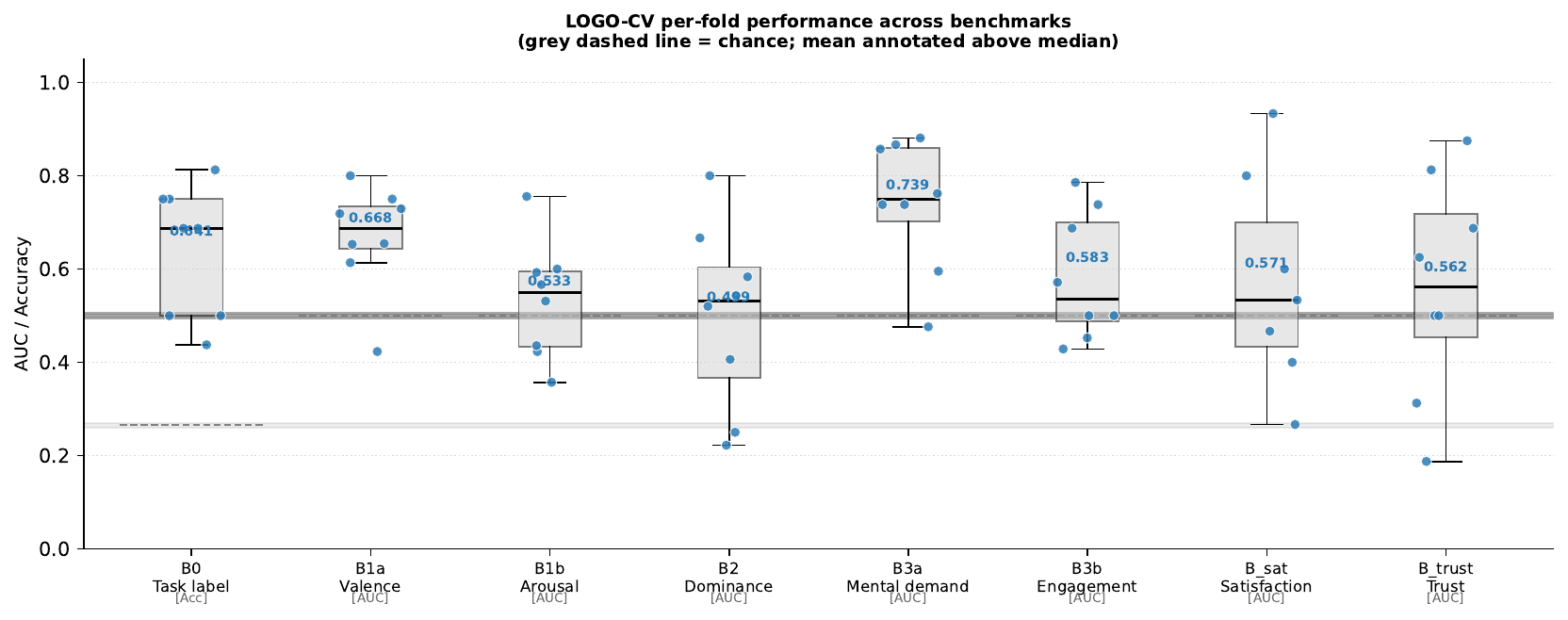}
  \caption{LOGO-CV per-fold performance strip plot across all benchmarks.
    Each dot is one fold; the box shows the interquartile range; the dashed
    grey line marks the chance baseline for each benchmark.
    Mean performance is annotated above the median.
    The wide fold variance for B5 (T4 contribution) and B4 (personality
    challenges) is visible directly from the fold distribution.
    Generated by \texttt{paper/analysis/feature\_importance.py}.}
  \label{fig:fold_strip}
\end{figure}

\section{Per-Session Quality Table}
\label{app:quality}

\Cref{tab:session_quality} reports per-session modality coverage for active
tasks T1--T4.
grp-09 had three EmotiBit sensors with partial data loss; grp-11 and grp-12
had PPG and EDA channels affected by device faults.
All sessions retained full eye-tracking row availability; pupil usability
was reduced in grp-07 and grp-11 due to calibration issues.

\input{tables/session_quality}

\section{Responsible AI and Croissant Metadata}
\label{app:rai}

NeurIPS 2026 E\&D requires Croissant metadata with Responsible AI fields
\cite{neurips2026ed,neurips2026hosting}.
\Cref{tab:rai_summary} summarises the fields in the release Croissant file.

\begin{table}[ht]
  \centering
  \caption{Responsible AI metadata summary for the no-video release.}
  \label{tab:rai_summary}
  \small
  \resizebox{\linewidth}{!}{%
  \begin{tabular}{@{}p{0.22\linewidth}p{0.34\linewidth}p{0.34\linewidth}@{}}
    \toprule
    \textbf{Croissant RAI field} & \textbf{Release statement} &
    \textbf{Reviewer-facing implication} \\
    \midrule
    \texttt{rai:\allowbreak{}dataLimitations}
      & Small, single-site, English-language cohort; no room video or
        room-frame gaze in v1; wearable HRV precision limited by PPG
        sampling.
      & Use for dataset characterisation and feasibility baselines, not
        population-general claims. \\
    \texttt{rai:dataBiases}
      & Convenience sample with likely university/community skew.
      & Do not treat group behaviours as culturally universal. \\
    \texttt{rai:personal\allowbreak{}Sensitive\allowbreak{}Information}
      & Age, sex, education, personality, physiology, egocentric gaze, and
        voice may be sensitive.
      & Identity mappings excluded; raw audio requires DUA. \\
    \texttt{rai:dataUseCases}
      & Strongest for task-state comparisons, modality QC, and exploratory
        group-affect modelling.
      & Not validated for clinical diagnosis or personnel evaluation. \\
    \texttt{rai:data\allowbreak{}SocialImpact}
      & Positive: transparent evaluation resource. Negative risks: affective
        surveillance and voice re-identification.
      & Mitigated by anonymisation, access tiers, and datasheet. \\
    \texttt{rai:has\allowbreak{}SyntheticData}
      & No synthetic data; derived from direct laboratory collection.
      & Provenance in collection, BIDS export, and benchmark scripts. \\
    \bottomrule
  \end{tabular}%
  }
\end{table}

\section{Datasheet for Datasets}
\label{app:datasheet}

This datasheet follows the Gebru et al.\ \cite{gebru2021datasheets} template.

\subsection*{Motivation}

\textbf{For what purpose was the dataset created?}
GroupAffect-4 was created to support research on multimodal group affect, social
dynamics, and interaction behaviour in structured co-located tasks.
Existing corpora focus on single-participant or dyadic settings; GroupAffect-4
fills a gap by providing synchronised physiology, eye tracking, audio, and
self-report for four-person groups.
\textbf{Who created the dataset?}
Meisam Jamshidi Seikavandi$^{1,2}$ (corresponding author), Alice Modica$^{1,3}$,
Anna Obara$^{1,4}$, Shan Ahmed Shaffi$^{1}$, Fabricio Batista Narcizo$^{1,2}$,
Tanya Ignatenko$^{1}$, Ted Vucurevich$^{1}$, Karim Haddad$^{1}$,
Daniel Barratt$^{3}$, Daniel Overholt$^{4}$, Jesper B{\"u}nsow Boldt$^{1}$,
Paolo Burelli$^{2}$, and Andrew Burke Dittberner$^{1}$.
$^{1}$GN Advanced Science, GN Group, Ballerup, Denmark;
$^{2}$IT University of Copenhagen, Copenhagen, Denmark;
$^{3}$Copenhagen Business School, Copenhagen, Denmark;
$^{4}$Aalborg University, Denmark.
Correspondence: \href{mailto:mjseikavandi@gn.com}{mjseikavandi@gn.com}.
\textbf{Who funded the creation?}
Internal research funding from GN Advanced Science, GN Group, Ballerup, Denmark.

\subsection*{Composition}

\textbf{What do the instances represent?}
Each instance is a participant-session-task recording unit.
The dataset contains 40 participants, 10 groups of 4, across 5 tasks
(T0 + T1--T4): up to 200 participant-task rows per modality.
\textbf{How many instances?}
200 expected participant-task recording units.
Coverage varies by modality (see \Cref{tab:modalities} and
\Cref{app:quality}).
\textbf{Does it contain all possible instances?}
This is the full paper-facing subset.
16 sessions were recorded in total: the first 5 were pilot sessions used to
finalise the protocol; of the 11 non-pilot sessions, 10 are released and 1
was excluded due to incomplete modality coverage.
\textbf{What data does each instance consist of?}
(1) EmotiBit physiological signals (PPG, EDA, skin temperature, IMU,
$\approx$25~Hz); (2) Tobii Pro Glasses~3 egocentric eye-tracking
($\approx$50~Hz); (3) close-talk lapel microphone audio (48~kHz WAV);
(4) transcript artifacts and turn-taking summaries derived from the audio;
(5) tablet self-report probes (valence, arousal); (6) task outcome records.
\textbf{Are there labels or targets?}
Self-reported valence and arousal (9-point SAM), task performance outcomes
where derivable, and BFI-44 personality traits (participant-level).
No external affect annotations.
\textbf{Is information missing?}
Yes.
EmotiBit availability missing for 9.0\% of expected rows;
PPG/EDA/temperature usability $\approx$24\% incomplete;
Tobii row availability 98.0\%;
valence/arousal self-report 14.5\% incomplete.
All missingness is documented in \Cref{app:quality}.
\textbf{Are relationships between instances explicit?}
Yes; participants within a session share a \texttt{group\_id}, and seat
identifiers P1--P4 are consistent across tasks.
\textbf{Are there errors, noise, or redundancies?}
Known artefacts: HRV RMSSD quantisation floor at $\approx$40~ms; audio
lapel misclassification of ambient noise in T0; eye-tracker calibration
drift in some sessions.
\textbf{Is the dataset self-contained?}
Self-contained for tabular modalities.
The public release includes transcript artifacts; room video, 3D
pose, and room-frame gaze are not in v1.
\textbf{Does it contain confidential data?}
Raw audio carries re-identification risk and is released under a separate
access tier.
\textbf{Does it relate to people?}
Yes; all data from human participants with written informed consent.
\textbf{Does it identify subpopulations?}
Self-reported age, sex, handedness, English proficiency, and education are
released to enable demographic reporting, not for subgroup targeting.
Racial or ethnic composition was not collected as part of the study protocol;
the sample is a university/community convenience sample and should not be
treated as representative of any demographic population.
\textbf{Is identification possible?}
Voice re-identification is a realistic risk; the DUA prohibits attempts.
\textbf{Does it contain sensitive data?}
Physiological signals combined with personality scores could support
inferential attacks; users are prohibited from applying data for clinical
diagnosis or individual profiling.

\subsection*{Collection Process}

\textbf{How was data acquired?}
Physiology via EmotiBit wrist sensors (LSL, $\approx$25~Hz); eye tracking
via Tobii Pro Glasses~3 ($\approx$50~Hz); audio via DPA~4060 lapel
microphones (48~kHz WAV); self-report via tablet-based SAM probes;
synchronisation via common LSL clock.
\textbf{What mechanisms were used?}
One or two experimenters per session; participants wore sensors throughout;
LSL markers demarcate task windows.
\textbf{Sampling strategy?}
Convenience sampling via internal staff networks and affiliated university
community postings at GN Advanced Science, GN Group, Ballerup, Denmark;
not nationally representative.
\textbf{Timeframe?}
March 2026.
\textbf{Ethical review?}
Yes; participant information statement and informed consent procedure reviewed
and approved by the GN Hearing A/S Legal Department; written informed consent
obtained from all participants; participants could withdraw at any time.

\subsection*{Preprocessing, Cleaning, and Labelling}

\textbf{Was preprocessing done?}
Yes: task-window extraction from LSL markers, per-modality QC passes,
BFI-44 domain scoring with reverse coding, and anonymisation.
\textbf{Was raw data saved?}
Yes; the full raw acquisition archive is retained internally.
The no-video release contains task-split BIDS-style modality files with
documented provenance.
\textbf{Is the software available?}
Yes; processing scripts under \texttt{tools/}; deterministic pipeline.

\subsection*{Uses}

\textbf{Has it been used for any tasks?}
Only by the authors for the analyses in this paper.
\textbf{What other tasks could it support?}
Multimodal affect modelling, group dynamics, personality--behaviour
associations, cross-modal synchrony, speech prosody and interaction rhythm.
See \Cref{sec:stats} and \Cref{sec:benchmarks}.
\textbf{Are there tasks it should not be used for?}
Clinical diagnosis, individual mental-health inference, workplace
surveillance, automated hiring, speaker re-identification, or voice cloning.

\subsection*{Distribution}

\textbf{How will it be distributed?}
The dataset is publicly available.
Code and processing scripts: \url{https://github.com/meisamjam/GroupAffect-4}.
Tabular features and derived data: \url{https://zenodo.org/records/20037847}
(persistent DOI, Zenodo, with Croissant metadata and datasheet).
Raw audio is available under a separate Data Use Agreement.
\textbf{How will the public release be distributed?}
Already publicly released via GitHub and Zenodo as described above.
\textbf{Under what licence?}
Tabular modalities (including audio-derived prosodic features): CC~BY~4.0,
consistent with participant informed consent.
Raw audio: Data Use Agreement (separate access tier).
\textbf{When?}
Released publicly at the time of this submission (v1.0).

\subsection*{Maintenance}

\textbf{Who will maintain the dataset?}
Meisam Jamshidi Seikavandi (\href{mailto:mjseikavandi@gn.com}{mjseikavandi@gn.com}),
with institutional support from GN Advanced Science, GN Group.
Minimum 5-year commitment via versioned tags at the Zenodo repository
(\url{https://zenodo.org/records/20037847}).
\textbf{Will it be updated?}
The dataset may be extended with video-derived 3D pose or room-frame gaze
after full QC, under new version tags.
This version is v1.0.
\textbf{What ethical review covered the data collection?}
Participant information statement and informed consent procedures reviewed
and approved by the GN Hearing A/S Legal Department;
written informed consent from all participants covering academic release of
all modalities, including audio-derived tabular features.

\newpage
\section*{NeurIPS Paper Checklist}


\begin{enumerate}

\item {\bf Claims}
\item[] Question: Do the main claims made in the abstract and introduction
  accurately reflect the paper's contributions and scope?
\item[] Answer: \answerYes{}
\item[] Justification: The abstract states that the paper introduces
  GroupAffect-4, a multimodal dataset of ten four-person collaborative
  sessions, and reports feasibility benchmarks across three analysis
  levels.  Both the dataset (\S\ref{sec:modalities}--\ref{sec:processing})
  and the benchmarks (\S\ref{sec:benchmarks}) are fully delivered and
  evaluated in the paper.
\item[] Guidelines:
  \begin{itemize}
    \item The answer NA means that the abstract and introduction do not
      include the claims made in the paper.
    \item The abstract and/or introduction should clearly state the claims
      made, including the contributions made in the paper and important
      assumptions and limitations.
    \item The claims made should match theoretical and experimental results,
      and reflect how much the results can be expected to generalise to
      other settings.
    \item It is fine to include aspirational goals as future work as long
      as it is clear that these are not currently demonstrated.
  \end{itemize}

\item {\bf Limitations}
\item[] Question: Does the paper discuss the limitations of the work?
\item[] Answer: \answerYes{}
\item[] Justification: A dedicated \S\ref{sec:limitations} discusses
  sample size ($n{=}10$ groups), single-site and English-language
  constraint, fixed task order, physiological signal missingness, and
  deferred modalities.  Extended caveats appear in
  \Cref{app:extended_limitations}.
\item[] Guidelines:
  \begin{itemize}
    \item The answer NA means that the paper has no limitations while the
      answer No means that the paper has limitations but does not discuss
      them.
    \item The authors are encouraged to create a separate ``Limitations''
      section in their paper.
    \item The paper should point out any strong assumptions and how robust
      the results are to violations of these assumptions (e.g., independence
      assumptions, additional assumptions to establish formal guarantees).
    \item The authors should reflect on how these limitations can affect
      the usability of the paper, e.g., in terms of generalisability of
      results and impact on future work.
  \end{itemize}

\item {\bf Theory Assumptions and Proofs}
\item[] Question: For each theoretical result, does the paper provide the
  full set of assumptions and a complete proof?
\item[] Answer: \answerNA{}
\item[] Justification: The paper makes no theoretical claims; it is a
  dataset and benchmark paper.
\item[] Guidelines:
  \begin{itemize}
    \item The answer NA means that the paper does not include theoretical
      results.
    \item All the theorems, formulas, and proofs in the paper should be
      numbered and cross-referenced.
    \item All assumptions should be clearly stated or referenced in the
      statement of any theorems.
    \item The proofs can either appear in the main paper or the
      supplemental material, but if they appear in the supplemental
      material, the authors are encouraged to provide a short proof sketch
      to provide intuition.
  \end{itemize}

\item {\bf Experimental Reproducibility}
\item[] Question: Does the paper fully disclose all the information needed
  to reproduce the main experimental results of the paper to the extent
  that it affects the main claims and is \emph{not} covered by the
  accompanying dataset paper?
\item[] Answer: \answerYes{}
\item[] Justification: The full processing pipeline and benchmark code are
  publicly released at \url{https://github.com/meisamjam/GroupAffect-4}.
  Preprocessing steps, feature selection, LOGO-CV split logic, and Ridge
  baseline hyperparameters are described in \S\ref{sec:processing},
  \S\ref{sec:benchmarks}, and \Cref{app:preprocessing,app:bench_notes}.
\item[] Guidelines:
  \begin{itemize}
    \item The answer NA means that the paper does not include experiments.
    \item If the paper includes experiments, a No answer to this question
      will not block the paper from being reviewed, but the authors are
      encouraged to include information needed to reproduce the results in
      their main paper.
    \item If the reproducibility of experimental results is crucial,
      provide a description of the code used to run the experiments.
  \end{itemize}

\item {\bf Open access to data and code}
\item[] Question: Does the paper provide open access to the data and code,
  with sufficient instructions to replicate the results, across all
  the experiments?
\item[] Answer: \answerYes{}
\item[] Justification: Tabular features and derived data are openly
  released on Zenodo (\url{https://zenodo.org/records/20037847}) under
  CC~BY~4.0; processing scripts and benchmark code are on GitHub
  (\url{https://github.com/meisamjam/GroupAffect-4}).
  Raw audio is available under a Data Use Agreement due to
  voice re-identification risk (\S\ref{sec:ethics}).
\item[] Guidelines:
  \begin{itemize}
    \item The answer NA means that paper does not include experiments
      requiring code.
    \item Please see the NeurIPS code and data submission guidelines
      (\url{https://nips.cc/public/guides/CodeSubmissionPolicy}) for more
      details.
    \item While we encourage the release of code and data, we understand
      that this might not be possible, so No is an acceptable answer.
      Authors of accepted papers will be asked to provide a link to a
      GitHub issues page for users to report bugs.
  \end{itemize}

\item {\bf Experimental Setting/Details}
\item[] Question: Does the paper specify all the training and test splits,
  evaluation metrics, model architectures, and hyperparameters?
\item[] Answer: \answerYes{}
\item[] Justification: \S\ref{sec:benchmarks} specifies Leave-One-Group-Out
  cross-validation (LOGO-CV), AUC, MAE, and Spearman $\rho$ as metrics.
  Baseline models are Ridge regression / logistic classifiers with
  default regularisation; no architecture search is performed.
  \Cref{app:bench_notes} provides per-benchmark interpretation and
  modality ablation details.
\item[] Guidelines:
  \begin{itemize}
    \item The answer NA means that the paper does not include experiments.
    \item The experimental setting should be presented in the core of the
      paper to a level of detail that is necessary to appreciate the
      results and make sense of them.
    \item The full details can be provided either with the code, in
      appendix, or both.
  \end{itemize}

\item {\bf Experiment Statistical Significance}
\item[] Question: Does the paper report error bars suitably and correctly,
  using standard deviations or confidence intervals?
\item[] Answer: \answerYes{}
\item[] Justification: Benchmark tables (\Cref{tab:benchmark_results} and
  \Cref{app:bench_notes}) report standard deviations across LOGO-CV folds.
  The paper notes where fold counts are too small for stable AUC estimates
  (Level-2 personality benchmarks, $n_{\text{fold}}{=}4$).
\item[] Guidelines:
  \begin{itemize}
    \item The answer NA means that the paper does not include experiments.
    \item The authors should answer ``Yes'' if the results are accompanied
      by error bars, confidence intervals, or statistical significance
      tests, at least for the experiments that support the main claims of
      the paper.
    \item If error bars are not included in the figures and tables, then
      the results reported should be accompanied by a description of how
      they were calculated.
  \end{itemize}

\item {\bf Experiments Compute Resources}
\item[] Question: For each experiment, does the paper provide sufficient
  information on the computer resources (type of compute workers, memory,
  time of execution) used to reproduce it?
\item[] Answer: \answerYes{}
\item[] Justification: All benchmarks use Ridge regression and logistic
  classifiers (scikit-learn defaults) on a standard CPU workstation.
  No GPU or specialised hardware is required; runtimes are seconds to
  minutes per fold.  The low-compute nature is noted in
  \S\ref{sec:benchmarks}.
\item[] Guidelines:
  \begin{itemize}
    \item The answer NA means that the paper does not include experiments.
    \item The paper should indicate the type of compute workers CPU/GPU,
      internal cluster, cloud provider, expected duration.
    \item The paper should provide enough information that a reader can
      estimate the total amount of compute and make sure their usage is
      consistent with NeurIPS' sustainability goal.
  \end{itemize}

\item {\bf Code Of Ethics}
\item[] Question: Does the research conform to the NeurIPS Code of Ethics?
\item[] Answer: \answerYes{}
\item[] Justification: Participants were recruited voluntarily, provided
  informed written consent, and were compensated.  No deception was
  employed.  Personal identifiers are not included in the public release.
  The dataset is gated against high-risk uses (\S\ref{sec:ethics}).
\item[] Guidelines:
  \begin{itemize}
    \item The answer NA means that the authors have not reviewed the
      NeurIPS Code of Ethics.
    \item If the authors answer No, they should explain the special
      circumstances that require a deviation from the Code of Ethics.
    \item The authors should make sure to adhere to its guidelines and
      send a petition to address the violation if needed.
  \end{itemize}

\item {\bf Broader Impacts}
\item[] Question: Does the paper discuss both potential positive societal
  impacts and negative societal impacts of the work performed?
\item[] Answer: \answerYes{}
\item[] Justification: \S\ref{sec:ethics} discusses misuse risks (clinical
  inference, workplace surveillance, speaker re-identification), access
  tiers to mitigate voice-data risks, and prohibited applications.
  Positive impacts (affective computing, assistive hearing technology,
  group-dynamics research) are discussed in \S\ref{sec:intro} and
  \S\ref{sec:discussion}.
\item[] Guidelines:
  \begin{itemize}
    \item The statement should include both positive and negative impacts.
    \item At minimum, a brief statement of potential negative impacts
      should be included in the paper (e.g., in the introduction or a
      dedicated section).
  \end{itemize}

\item {\bf Safeguards}
\item[] Question: Does the paper describe safeguards that have been put
  in place for responsible release of data or models that have a high risk
  of being misused (e.g., pretrained language models, image generators, or
  scraped datasets)?
\item[] Answer: \answerYes{}
\item[] Justification: Raw audio is gated behind a Data Use Agreement to
  limit voice re-identification risk.  The datasheet (\Cref{app:datasheet})
  explicitly lists prohibited uses.  The public tabular release excludes
  all direct participant identifiers.
\item[] Guidelines:
  \begin{itemize}
    \item Datasets that can be used to train models with potential for
      significant harm should be released with necessary safeguards, for
      example, only on a case-by-case basis or not at all.
    \item NeurIPS expects that the authors describe the safeguards in
      place and any limitations of those safeguards.
  \end{itemize}

\item {\bf Licenses for Existing Assets}
\item[] Question: Are the creators or original owners of assets (e.g.,
  code, data, models), used in this paper, properly credited and given
  an appropriate citation?
\item[] Answer: \answerYes{}
\item[] Justification: All third-party tools and libraries are cited:
  EmotiBit SDK, Tobii Pro SDK, openSMILE, scikit-learn, pandas, numpy,
  and associated works (\S\ref{sec:modalities}, \S\ref{sec:processing}).
\item[] Guidelines:
  \begin{itemize}
    \item Regardless of whether you created or used third-party assets,
      ensure that you cite the relevant paper(s).
    \item When using existing assets, the relevant licenses must be
      respected.
    \item If you created a new asset (e.g., a new dataset, new model, new
      code), list the license of the new asset in the paper.
  \end{itemize}

\item {\bf New Assets}
\item[] Question: Are new assets introduced in the paper well documented
  and is the documentation provided alongside the assets?
\item[] Answer: \answerYes{}
\item[] Justification: The GroupAffect-4 dataset is documented via a
  Datasheets for Datasets section (\Cref{app:datasheet}), Croissant
  metadata file (included in the Zenodo deposit), and README in the
  GitHub repository.  Licenses (CC~BY~4.0 for tabular data; separate
  DUA for raw audio) are stated in \S\ref{sec:ethics} and the datasheet.
\item[] Guidelines:
  \begin{itemize}
    \item Upload the data, code, and model to an anonymised website such
      as Anonymous GitHub or Zenodo in a structured way so that it is easy
      for a reviewer to understand.
    \item Code should be organized and have a README.
    \item Models weights should have a model card.
    \item New datasets should be accompanied by a datasheet.
  \end{itemize}

\item {\bf Crowdsourcing and Research with Human Subjects}
\item[] Question: For crowdsourcing experiments and research with human
  subjects, does the paper include the following information: (i) the
  full text or a link to informed consent forms, (ii) instructions given
  to the participants and (iii) compensation?
\item[] Answer: \answerYes{}
\item[] Justification: \S\ref{sec:design} and the datasheet
  (\Cref{app:datasheet}) describe participant recruitment, session
  procedures, and compensation.  Participants were recruited via staff
  networks and affiliated university postings; sessions were conducted
  with written informed consent.  The participant information statement
  is included in the public Zenodo release (\url{https://zenodo.org/records/20037847}).
\item[] Guidelines:
  \begin{itemize}
    \item Ideally, this information should be provided as supplemental
      material or as a link to a URL containing this information.
  \end{itemize}

\item {\bf Institutional Review Board (IRB) Approvals or Equivalent for
  Research with Human Subjects}
\item[] Question: Does the paper describe whether Institutional Review
  Board (IRB) approvals or equivalent were obtained?
\item[] Answer: \answerYes{}
\item[] Justification: The participant information statement and informed
  consent procedures were reviewed and approved by the GN Hearing A/S
  Legal Department, as stated in \S\ref{sec:ethics} and the datasheet
  maintenance section (\Cref{app:datasheet}).  Written informed consent
  was obtained from all participants covering academic release of all
  modalities.
\item[] Guidelines:
  \begin{itemize}
    \item Depending on the country in which research was conducted, IRB
      approval (or equivalent) may be required for any human subjects
      research.
    \item If the authors obtained IRB approval, they should clearly state
      this in the paper.
    \item We recognize that the procedures for this may vary significantly
      between institutions and locations, and we expect authors to adhere
      to the NeurIPS Code of Ethics and the guidelines for their
      institution.
    \item For initial submissions, do not include any information that
      would break anonymity (if applicable), such as the institution
      conducting the review.
  \end{itemize}

\end{enumerate}

\end{document}

%% file: tables/task_overview.tex

\begin{table}[t]
  \centering
  \caption{Session structure: tasks with phases' timer durations as displayed on the shared screen.}
  \label{tab:tasks}
  \small
  \begin{tabular}{ll>{\centering\arraybackslash}p{0.3\linewidth}}
    \toprule
    \textbf{ID} & \textbf{Task} & \textbf{Duration (s)} \\
    \midrule
    T0 & Free-talk baseline
       & 300 (free talk) \\
    T1 & Hidden-profile decision
       & 75 (reading); 420 (discussion); 60 (selection) \\
    T2 & Mini-negotiation
       & 480 (negotiation); no timer (settlement form) \\
    T3 & Idea generation and selection
       & 180 (generation); 420 (discussion); 60 (selection) \\
    T4 & Public-goods micro-game
       & 60 (contribution); 60 (reveal); 180 (discussion) \\
    \bottomrule
  \end{tabular}
\end{table}

%% file: tables/benchmark_results.tex
\begin{table*}[t]
  \centering
  \caption{Benchmark definitions and leave-one-group-out results (31-feature leakage-corrected set).
    Fold-mean performance with across-fold SD and bootstrap 95\% CI (5000 iterations).
    Chance baselines: majority-class Acc.\ (B0), AUC~0.500 (classification), naive-mean MAE (B6/B7).
    Modalities: Ph\,=\,physiology, Pu\,=\,pupil, Au\,=\,audio; \textit{All}\,=\,Ph+Pu+Au+BFI.
    $\dagger$~new addition; [chall.]~challenge benchmark; $\ddag$~CI excludes chance.
    B0--B3d estimates are biased (pre-split normalisation; see \Cref{sec:processing}).}
  \label{tab:benchmark_results}
  \phantomsection\label{tab:benchmark_defs}
  \small
  \setlength{\tabcolsep}{3pt}
  \begin{tabular}{lp{0.26\linewidth}rrlll}
    \toprule
    \textbf{ID} & \textbf{Target} & $\bm{n}$ & \textbf{Metric} & \textbf{Mod.} & \textbf{Mean (SD)} & \textbf{95\% CI} \\
    \midrule
    \multicolumn{7}{@{}l}{\textit{Sanity check (Pt-task unit)}} \\
    B0 & Task label (T1--T4)                             & 136 & Acc.  & Ph+Pu+Au & 0.641 (0.132) & {[0.55, 0.73]} \\
    \midrule
    \multicolumn{7}{@{}l}{\textit{Level 1 --- Within-person state (Pt-task unit)}} \\
    B1a & Valence (hi/lo)                                & 107 & AUC   & All      & 0.657 (0.110) & {[0.58, 0.73]} \\
    B1b & Arousal (hi/lo)                                & 107 & AUC   & All      & 0.528 (0.114) & {[0.45, 0.60]} \\
    B2  & Dominance (hi/lo)                              &  83 & AUC   & Ph+Pu+Au & 0.499 (0.186) & {[0.37, 0.62]} \\
    B3a & Mental demand (hi/lo)                          &  99 & AUC   & All      & 0.719 (0.142) & {[0.62, 0.81]} \\
    B3b & Engagement (hi/lo)                             &  99 & AUC   & All      & 0.591 (0.136) & {[0.50, 0.69]} \\
    B3c & Satisfaction (T2--T3)$^\dagger$                &  60 & AUC   & All      & 0.571 (0.213) & {[0.419, 0.733]} \\
    B3d & Trust pooled (T2, T4)$^\dagger$                &  60 & AUC   & All      & 0.562 (0.221) & {[0.406, 0.711]} \\
    \quad & B3d T4-only (corrected)$^\dagger$            &  28 & AUC   &          & 0.679 (0.220) & {[0.536, 0.857]} \\
    \midrule
    \multicolumn{7}{@{}l}{\textit{Level 2 --- Between-person traits (Participant unit, CHALLENGE)}} \\
    B4a & BFI Extraversion$^\ddag$                       &  32 & AUC   & Pt-mean all & 0.396 (0.353) & {[0.17, 0.65]} \\
    B4b & BFI Openness$^\ddag$                           &  32 & AUC   & Pt-mean all & 0.306 (0.244) & {[0.11, 0.50]} \\
    B4c & BFI Agreeableness$^\dagger$                    &  31 & AUC   & Pt-mean all & 0.604 (0.424) & {[0.292, 0.875]} \\
    \quad & B4c (top-2 T2 feats$^\dagger$    &  31 & AUC   &          & 0.625 (0.317) & {[0.406, 0.844]} \\
    B5  & T4 contribution (median split)                 &  28 & AUC   & Pt-mean all & 0.429 (0.290) & {[0.21, 0.64]} \\
    \midrule
    \multicolumn{7}{@{}l}{\textit{Level 3 --- Group dynamics (Grp-task unit)}} \\
    B6a & Speaking Gini (Mean)           &  38 & MAE   & Ph+Pu+BFI & 0.089 & --- \\
    B6b & Speaking Gini (SD )$^\dagger$$^\ddag$  &  28 & MAE   & Ph+Au     & 0.102 (0.053) & {[0.068, 0.140]} \\
    \quad & B6b binary (SD$_\text{raw}$)      &  28 & AUC   &          & 0.952 (0.117) & {[0.857, 1.000]} \\
    B7  & Speech-overlap fraction                        &  38 & MAE   & Ph+Pu+BFI & 0.063 & --- \\
    \bottomrule
  \end{tabular}
\end{table*}

%% file: tables/dataset_stats.tex

\begin{table}[t]
  \centering
  \caption{Dataset statistics and task-response completeness.
    QC pass: 10 sessions, 200 expected participant-task rows.
    Signal usability per channel (EmotiBit: $\geq$80\% finite samples; Tobii gaze: $\geq$70\% valid).}
  \label{tab:stats}
  \phantomsection\label{tab:task_response_validity}
  \small
  \begin{tabular}{lr}
    \toprule
    \textbf{Metric} & \textbf{Value} \\
    \midrule
    Groups / unique participants          & 10 / 40 \\
    Age, mean (SD), range                 & 35.4~(9.2), 23--58 \\
    Sex                                   & 21F / 19M \\
    Scheduled recording time              & $\sim$11.7~h \\
    \midrule
    \multicolumn{2}{l}{\textit{Signal QC (200 expected rows)}} \\
    \midrule
    EmotiBit rows observed                & 182/200~(91.0\%) \\
    \quad PPG / EDA / Temp.\ usable       & 76.0\% / 75.5\% / 76.0\% \\
    \quad IMU usable                      & 181/200~(90.5\%) \\
    Tobii rows observed                   & 196/200~(98.0\%) \\
    \quad Gaze / Pupil usable             & 87.2\% / 95.4\% \\
    \midrule
    \multicolumn{2}{l}{\textit{Task-response completeness (all 10 groups completed all 4 tasks)}} \\
    \midrule
    \ac{VAD} rows (T1/T2/T3/T4)               & 34 / 36 / 38 / 35 of 40 \\
    Core post-task items (T1/T2/T3/T4)   & 712/712 / 590/600 / 633/640 / 319/320 \\
    Overall post-task completeness        & 2254/2280~(98.9\%) \\
    T1 modal decision / T2 settlement     & 10/10 / 10/10 \\
    T3 winning-idea records               & 7/10 groups \\
    \bottomrule
  \end{tabular}
\end{table}

%% file: tables/modalities.tex

\begin{table}[t]
  \centering
  \caption{Modalities released in this version of GroupAffect-4.
    All per-participant streams use seat identifiers P1--P4.
    Sampling rates marked with $\dagger$ are channel-dependent;
    see the per-channel table in the appendix.
    Transcript artifacts are included in the reviewer-access package.
    Multi-camera video and marker-assisted 3D outputs were recorded or piloted
    for follow-up work; they are not part of the current release or analysis.}
  \label{tab:modalities}
  \footnotesize
  \setlength{\tabcolsep}{4pt}
  \begin{tabular}{@{}p{0.16\linewidth}p{0.20\linewidth}p{0.31\linewidth}p{0.11\linewidth}p{0.14\linewidth}@{}}
    \toprule
    \textbf{Modality} & \textbf{Device} & \textbf{Signals} & \textbf{Rate} & \textbf{Per session} \\
    \midrule
    Physiology
      & EmotiBit
      & PPG (3 wavelengths), EDA, temperature, IMU
      & $\dagger$
      & 4 (one per P) \\
    \addlinespace
    Eye tracking (egocentric)
      & Tobii Pro Glasses 3
      & Head-relative scene-frame gaze, pupil diameter, validity
      & 50~Hz
      & 4 (one per P) \\
    \addlinespace
    Audio (close-talk)
      & DPA 4060
      & Mono, 48~kHz / 16-bit
      & 48~kHz
      & 4 (one per P) \\
    \addlinespace
    Audio (room)
      & DPA 4060
      & Mono, 48~kHz / 16-bit
      & 48~kHz
      & 1 \\
    \addlinespace
    Transcript artifacts
      & Audio post-processing
      & Transcript and turn-taking tables derived from participant audio
      & event-driven
      & 4 participant channels \\
    \addlinespace
    Behavioural self-report
      & Android tablets
      & \ac{VAD} probes, task responses
      & event-driven
      & 4 (one per P) \\
    \addlinespace
    Personality (pre-session)
      & Online questionnaire
      & BFI-44 item responses, demographics
      & one-shot
      & 4 (one per P) \\
    \bottomrule
  \end{tabular}
\end{table}

%% file: tables/preprocessing_steps.tex
\begin{table}[t]
  \centering
  \caption{Step-by-step preprocessing audit across 136 active participant-task rows and
    102 candidate feature columns (13,872 feature$\times$row cells).
    Values \emph{gated} are replaced with NaN; values \emph{clipped} are winsorised in-place;
    values \emph{transformed} are rescaled (within-person z-score); the KNN row
    is a descriptive audit, while benchmark imputation is fit inside each LOGO-CV fold.
    After global feature selection (missing $>$50\% and $|r|>0.95$ greedy removal)
    35 features survive for benchmark evaluation (5 biomarker composites excluded; see text).}
  \label{tab:preprocessing}
  \small
  \setlength{\tabcolsep}{5pt}
  \begin{tabular}{lrrrr}
    \toprule
    \textbf{Step} & \textbf{New NaNs} & \textbf{Clipped} & \textbf{Transformed} & \textbf{Imputed} \\
    \midrule
    ET quality gating$^\dagger$   &  16 &   0 &     0 &   0 \\
    Plausibility gating$^\ddagger$ &  67 &   0 &     0 &   0 \\
    Winsorisation ($\pm3\sigma$)  &   0 &  32 &     0 &   0 \\
    Within-person z-score         &   0 &   0 & 9{,}970 &   0 \\
    KNN imputation audit ($k=5$)  &   0 &   0 &     0 & 772 \\
    \bottomrule
  \end{tabular}
  \vspace{2pt}
  {\footnotesize
    $^\dagger$Pupil features nulled when $>$50\% samples missing per row;
    gaze features nulled when valid fraction $<$50\%.\\
    $^\ddagger$Bounds: HR $\in$[40,180]\,bpm; HRV RMSSD $\in$[10,300]\,ms;
    EDA $\in$[0,25]\,$\mu$S; skin temp $\in$[28,40]\,°C; pupil $\in$[1.5,9.0]\,mm;
    pitch $\in$[5,55]\,semitones; $\Delta$HR $\in$[$-$40,+40]\,bpm; $\Delta$pupil $\in$[$-$3,+3]\,mm.
    The benchmark-ready table retains residual NaNs for fold-local KNN imputation.}
\end{table}

%% file: tables/autonomic_worked_example.tex

\begin{table}[t]
  \centering
  \caption{Descriptive task-level autonomic findings used in the worked
    example. Values are baseline-normalised against T0 within participant.
    Effect size is within-participant Cohen's $d_z$ against zero; rows are
    descriptive and are not treated as confirmatory tests.}
  \label{tab:autonomic_worked_example}
  \small
  \begin{tabular}{llrrr}
    \toprule
    \textbf{Task} & \textbf{Feature} & \textbf{$n$} & \textbf{Mean change} & \textbf{$d_z$} \\
    \midrule
    T1 & Pupil diameter (mm)          & 40 & -0.117 & -1.01 \\
    T2 & Pupil diameter (mm)          & 40 & -0.038 & -0.31 \\
    T3 & Pupil diameter (mm)          & 40 & -0.108 & -0.70 \\
    T4 & Pupil diameter (mm)          & 40 & -0.076 & -0.73 \\
    T1 & Skin temperature (deg C)     & 34 &  0.610 &  0.96 \\
    T2 & Skin temperature (deg C)     & 34 &  0.613 &  0.80 \\
    T3 & Skin temperature (deg C)     & 33 &  0.468 &  0.58 \\
    T4 & Skin temperature (deg C)     & 33 &  0.459 &  0.60 \\
    T1 & SCR rate (peaks/s)           & 33 &  0.027 &  0.51 \\
    T2 & SCR rate (peaks/s)           & 33 &  0.037 &  0.42 \\
    T3 & SCR rate (peaks/s)           & 32 &  0.036 &  0.39 \\
    T4 & SCR rate (peaks/s)           & 32 &  0.045 &  0.45 \\
    \bottomrule
  \end{tabular}
\end{table}

%% file: tables/construct_task_coverage.tex

\begin{table}[h]
  \centering
  \caption{Post-block survey construct availability by task.
    A tick (\checkmark) indicates the item appeared in the post-block
    questionnaire for that task; a dash indicates it was absent from that
    task's survey module.
    T0 had no post-block survey.
    Real-time \ac{VAD} probes (valence, arousal, dominance) were administered
    during all five tasks; dominance was absent from the T4 \ac{VAD} probe schema.}
  \label{tab:construct_task_coverage}
  \small
  \setlength{\tabcolsep}{5pt}
  \begin{tabular}{@{}lcccc@{}}
    \toprule
    \textbf{Construct} & \textbf{T1} & \textbf{T2} & \textbf{T3} & \textbf{T4} \\
    \midrule
    \multicolumn{5}{@{}l}{\textit{Shared core items}} \\
    Engagement          & \checkmark & \checkmark & \checkmark & \checkmark \\
    Mental demand       & \checkmark & \checkmark & \checkmark & \checkmark \\
    Overall valence     & \checkmark & \checkmark & \checkmark & \checkmark \\
    Per-seat dominance rating & \checkmark & \checkmark & \checkmark & \checkmark \\
    \addlinespace
    \multicolumn{5}{@{}l}{\textit{Partially shared items}} \\
    Voice \& inclusion  & \checkmark & \checkmark & \checkmark & --- \\
    Team coordination   & \checkmark & ---        & \checkmark & --- \\
    Satisfaction        & ---        & \checkmark & \checkmark & --- \\
    Trust (3 items)     & ---        & \checkmark & ---        & \checkmark \\
    Fairness            & ---        & ---        & \checkmark & \checkmark \\
    Confidence          & ---        & ---        & \checkmark & \checkmark \\
    Perceived control   & \checkmark & \checkmark & ---        & --- \\
    \addlinespace
    \multicolumn{5}{@{}l}{\textit{Task-specific items}} \\
    Decision confidence (T1)         & \checkmark & --- & --- & --- \\
    Info sharing (T1)                & \checkmark & --- & --- & --- \\
    Equality of contribution (T1)    & \checkmark & --- & --- & --- \\
    Familiarity per seat (T1)        & \checkmark & --- & --- & --- \\
    Cooperativeness (T2)             & ---        & \checkmark & --- & --- \\
    Idea quality / diversity (T3)    & ---        & --- & \checkmark & --- \\
    Psychological safety (T3)        & ---        & --- & \checkmark & --- \\
    Contribution tokens (T4)         & ---        & --- & --- & \checkmark \\
    Regret / social concern (T4)     & ---        & --- & --- & \checkmark \\
    Expectation match (T4)           & ---        & --- & --- & \checkmark \\
    \bottomrule
  \end{tabular}
\end{table}

%% file: tables/audio_turn_taking_summary.tex
\begin{table}[t]
  \centering
  \caption{Transcript-derived turn-taking and overlap summary by task (participants only).}
  \label{tab:audio_turn_taking_summary}
  \small
  \begin{tabular}{lrrrrr}
    \toprule
    \textbf{Task} & \textbf{Turns} & \textbf{Turn dur. (s)} & \textbf{Overlap frac.} & \textbf{Resp. gap (s)} & \textbf{Backchannel rate} \\ 
    \midrule
    T0 & 54.8 & 2.19 & 0.130 & 3.63 & 0.404 \\
    T1 & 21.0 & 4.41 & 0.125 & 3.60 & 0.305 \\
    T2 & 30.2 & 4.19 & 0.107 & 3.04 & 0.209 \\
    T3 & 27.4 & 3.16 & 0.191 & 5.19 & 0.271 \\
    T4 & 16.3 & 2.44 & 0.200 & 3.21 & 0.313 \\
    \bottomrule
  \end{tabular}
\end{table}

%% file: tables/ablation_table.tex
\begin{table}[t]
  \centering
  \caption{Feature-modality ablation: LOGO-CV primary metric under ten
    feature-subset conditions.
    \textbf{Cardiac}: HR mean and HRV RMSSD (delta + absolute).
    \textbf{EDA/GSR}: EDA tonic/phasic and skin temperature.
    \textbf{Pupil}: pupil dilation delta and absolute.
    \textbf{Audio}: speaking fraction, overlap, energy, pitch, prosody.
    \textbf{BFI}: Big Five personality traits.
    \textbf{Sensor}: all four sensor modalities combined (excludes annotation features).
    Bold = best condition per row.
    B6 (Speaking Gini) and B7 (Speech-overlap fraction): group-task level; physio/pupil are group means.
    $^\star$\,``B7 (legacy): Floor dominance'' is a binary classification target from an earlier benchmark version, retained here for ablation completeness; it does not appear in the main evaluation suite.
    \emph{Note:} This ablation uses accuracy for binary classification targets
    (B0, B2, B3a/b) and MAE for continuous/regression targets (B1a/b, B4a/b, B5),
    and was computed with the full 40-feature set (including biomarker composites
    subsequently excluded from the main benchmarks); results are therefore not
    directly comparable to the AUC figures in \Cref{tab:benchmark_results}.
    Audio dominates task classification; pupil and audio jointly dominate
    cognitive-state detection; notably, a single audio feature
    (\texttt{audio\_overlap\_fraction\_x}) matches or exceeds the full-sensor
    model for mental demand (see text).}
  \label{tab:ablation}
  \scriptsize
  \setlength{\tabcolsep}{3pt}
  \resizebox{\linewidth}{!}{%
  \begin{tabular}{llrrrrrrrrrr}
    \toprule
    \textbf{Benchmark} & \textbf{Metric} & \textbf{Cardiac} & \textbf{EDA/GSR} & \textbf{Pupil} & \textbf{Audio} & \textbf{BFI} & \textbf{Card+EDA} & \textbf{Card+Pup} & \textbf{EDA+Pup} & \textbf{Sensor} & \textbf{All+BFI} \\
    \midrule
    B0: Task label (T1-T & Acc. & 0.265 & 0.235 & 0.493 & 0.610 & -- & 0.301 & 0.515 & 0.500 & \textbf{0.640} & 0.581 \\
    \addlinespace[2pt]
    B1a: Valence & MAE & 1.462 & \textbf{1.030} & 1.094 & 1.118 & 1.070 & 1.640 & 1.148 & 1.087 & 1.442 & 1.725 \\
    B1b: Arousal & MAE & 1.147 & 1.105 & \textbf{1.077} & 1.159 & 1.121 & 1.195 & 1.103 & 1.102 & 1.344 & 1.330 \\
    \addlinespace[2pt]
    B2: Dominance (high/ & Acc. & 0.518 & 0.482 & 0.470 & 0.578 & 0.470 & 0.482 & 0.518 & 0.422 & \textbf{0.590} & 0.566 \\
    \addlinespace[2pt]
    B3b: Engagement & Acc. & 0.525 & 0.525 & 0.616 & 0.556 & 0.455 & 0.545 & 0.646 & 0.626 & \textbf{0.677} & 0.657 \\
    B3a: Mental demand & Acc. & 0.556 & 0.465 & 0.677 & \textbf{0.707} & 0.515 & 0.556 & 0.616 & 0.646 & 0.687 & 0.626 \\
    \addlinespace[2pt]
    B4a: BFI Extraversion & MAE & 0.508 & \textbf{0.473} & 0.608 & 0.644 & -- & 0.678 & 0.950 & 0.511 & 1.040 & -- \\
    B4b: BFI Openness & MAE & 0.960 & 0.476 & \textbf{0.464} & 0.652 & -- & 0.962 & 1.014 & 0.544 & 1.988 & -- \\
    \addlinespace[2pt]
    B5: T4 Contribution & MAE & 0.715 & 0.506 & 0.507 & 0.767 & \textbf{0.496} & 0.710 & 0.745 & 0.522 & 0.845 & 0.708 \\
    \addlinespace[2pt]
    B7 (legacy): Floor dominance$^\star$ & Acc. & 0.589 & 0.652 & 0.554 & -- & 0.554 & \textbf{0.670} & 0.446 & 0.580 & 0.562 & 0.598 \\
    \addlinespace[2pt]
    B6a: Speaking Gini & MAE & \textbf{0.083} & 0.094 & 0.097 & -- & 0.099 & 0.107 & -- & -- & 0.129 & 0.154 \\
    \addlinespace[2pt]
    B7: Speech-overlap fraction & MAE & 0.070 & 0.077 & \textbf{0.057} & 0.078 & 0.068 & 0.070 & -- & -- & 0.097 & 0.117 \\
    \bottomrule
  \end{tabular}%
  }
\end{table}

%% file: tables/session_quality.tex
\begin{table}[htbp]
\centering
\caption{Per-session modality coverage across the 10 final groups (active tasks T1--T4; 16 participant-task rows expected per session per modality). \checkmark\ = all 16/16 rows available/usable; fractions indicate partial coverage. Audio counts include T0--T4 WAV files across all microphone channels. Beh = behavioural/self-report event files.}
\label{tab:session_quality}
\small
\setlength{\tabcolsep}{4pt}
\begin{tabular}{llcccccc}
\toprule
Group & Date & Physio & PPG & EDA & ET & Pupil & Audio \\
      &      & avail. & usable & usable & avail. & usable & (WAV) \\
\midrule
grp-07 & 2026-03-12 & \checkmark & \checkmark & \checkmark & \checkmark & 12/16 & 30 \\
grp-08 & 2026-03-13 & \checkmark & \checkmark & \checkmark & \checkmark & \checkmark & 28 \\
grp-09 & 2026-03-17 & 8/16       & 7/16       & 6/16       & \checkmark & \checkmark & 30 \\
grp-10 & 2026-03-17 & \checkmark & \checkmark & \checkmark & \checkmark & \checkmark & 30 \\
grp-11 & 2026-03-18 & \checkmark & 4/16       & 4/16       & \checkmark & 12/16 & 30 \\
grp-12 & 2026-03-18 & 14/16      & 0/16       & 0/16       & \checkmark & \checkmark & 30 \\
grp-13 & 2026-03-18 & \checkmark & \checkmark & \checkmark & \checkmark & \checkmark & 22 \\
grp-14 & 2026-03-19 & \checkmark & \checkmark & \checkmark & \checkmark & \checkmark & 30 \\
grp-15 & 2026-03-19 & \checkmark & \checkmark & \checkmark & \checkmark & \checkmark & 34 \\
grp-16 & 2026-03-20 & \checkmark & \checkmark & \checkmark & \checkmark & \checkmark & 30 \\
\midrule
\textbf{Total} & & 150/160 & 123/160 & 122/160 & 160/160 & 152/160 & 294 \\
\textbf{(\%)} & & (93.8\%) & (76.9\%) & (76.2\%) & (100\%) & (95.0\%) & -- \\
\bottomrule
\end{tabular}
\end{table}